\documentclass[10pt,twocolumn,letterpaper]{article}

\usepackage{cvpr} 

\usepackage{amsmath,amsfonts,bm}

\def\eqref#1{equation~\ref{#1}}

\def\1{\bm{1}}

\DeclareMathAlphabet{\mathsfit}{\encodingdefault}{\sfdefault}{m}{sl}
\SetMathAlphabet{\mathsfit}{bold}{\encodingdefault}{\sfdefault}{bx}{n}

\usepackage[table, dvipsnames]{xcolor}
\usepackage[utf8]{inputenc} 
\usepackage[T1]{fontenc}    
\usepackage{url}            
\usepackage{booktabs}       
\usepackage{amsfonts}       
\usepackage{nicefrac}       
\usepackage{amsmath}
\usepackage{amssymb}
\usepackage{float}
\usepackage{microtype}      
\usepackage{acronym}
\usepackage{marvosym}
\usepackage{multirow}
\usepackage{pifont}
\usepackage{xspace}
\usepackage{graphicx}       
\usepackage{subcaption}     
\usepackage{wrapfig}
\definecolor{deeppink}{RGB}{180,20,140}

\definecolor{cvprblue}{rgb}{0.21,0.49,0.74}
\usepackage[pagebackref,breaklinks,colorlinks,allcolors=cvprblue]{hyperref}

\def\paperID{34348} 
\def\confName{CVPR}
\def\confYear{2026}

\title{Electromagnetic Inverse Scattering from a Single Transmitter}

\author{Yizhe Cheng\textsuperscript{1,}$^{*}$
\quad Chunxun Tian\textsuperscript{1,}$^{*}$ \quad Haoru Wang\textsuperscript{1} \quad
Wentao Zhu\textsuperscript{2} \quad Xiaoxuan Ma\textsuperscript{1,3\Letter} \quad Yizhou Wang\textsuperscript{1} \\
\textsuperscript{1~}Center on Frontiers of Computing Studies,
School of Computer Science, Peking University \\
\textsuperscript{2~}Institute of Digital Twin, Eastern Institute of Technology, Ningbo \\
\textsuperscript{3~}Robotics Institute, Carnegie Mellon University \\
{\tt\small $^{*}$Equal contribution, random order. $^{\textrm{\Letter}}$ Corresponding author.}
}

\begin{document}

\input{notation}

\twocolumn[{%
    \renewcommand\twocolumn[1][]{#1}
    \setlength{\tabcolsep}{0.0mm} 
    \newcommand{\sz}{0.125}  
    \maketitle
    \vspace{-7ex}
    \begin{center}
        \newcommand{\teaserwidth}{\textwidth}
        \includegraphics[width=\linewidth]{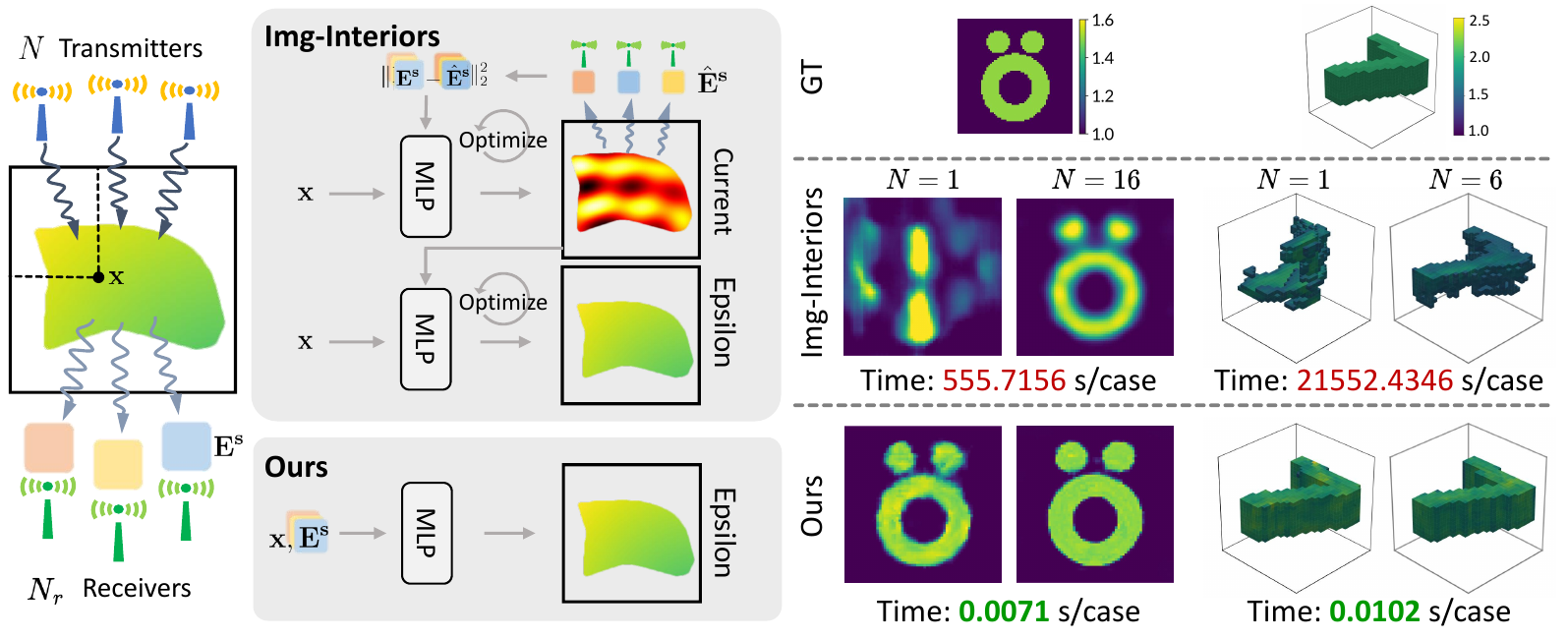}
    \vspace{-4ex}
    \captionof{figure}{\textbf{Comparison between our method and the previous state-of-the-art.} \textbf{Left:} \iishort \citep{luo2024imaging} requires case-specific optimization to reconstruct the permittivity. In contrast, our method is a data-driven framework that operates in an \textit{end-to-end}, \textit{feed-forward} manner for solving inverse scattering. \textbf{Right:} Our method yields more accurate reconstructions than \iishort \citep{luo2024imaging}. It remains robust even with a \textit{single} transmitter and achieves real-time inference with over $70,000\times$ speed-up. } 
    \label{fig:teaser}
    \end{center}
}]

\begin{abstract}
Electromagnetic Inverse Scattering Problems (EISP) seek to reconstruct relative permittivity from scattered fields and are fundamental to applications like medical imaging.
This inverse process is inherently ill-posed and highly nonlinear, making it particularly challenging, especially under sparse transmitter setups, \eg, with only one transmitter.
While recent machine learning-based approaches have shown promising results, they often rely on time-consuming, case-specific optimization and perform poorly under sparse transmitter setups.
To address these limitations, we revisit EISP from a data-driven perspective. The scarcity of transmitters leads to an insufficient amount of measured data, which fails to capture adequate physical information for stable inversion.
Accordingly, we propose a \textit{fully end-to-end} and \textit{data-driven} framework that predicts the relative permittivity of scatterers from measured fields, leveraging data distribution priors to compensate for the incomplete information from sparse measurements.
This design enables data-driven training and \textit{feed-forward} prediction of relative permittivity while maintaining strong robustness to transmitter sparsity.
Extensive experiments show that our method outperforms state-of-the-art approaches in reconstruction accuracy and robustness. 
Notably, we demonstrate, for the first time, high-quality reconstruction from a single transmitter.
This work advances practical electromagnetic imaging by providing a new, cost-effective paradigm to inverse scattering. 
Code and models are released at \url{https://gomenei.github.io/SingleTX-EISP/}.
\end{abstract}
\section{Introduction}

\label{sec:intro}

Electromagnetic waves can penetrate object surfaces, making them essential for non-invasive imaging~\citep{geng2023dreampcddeepreconstructionenhancement,o2018microwave}. At the core of electromagnetic imaging lies the \ac{eisp}, which seeks to reconstruct an object’s relative permittivity from measured scattered electromagnetic field~\citep{nikolova2011microwave}. By solving \ac{eisp}, we can accurately recover internal structures without physical intrusion~\citep{Song2005ThroughwallI}, enabling a range of scientific and industrial applications, such as safer and more cost-effective alternatives to X-rays and MRI scans~\citep{Bevacqua2021MillimeterWavesBC,o2018microwave,nikolova2011microwave}.
Typically, \ac{eisp} necessitate a large number of transmitters and receivers to acquire sufficient measurement data. This requirement, however, leads to increased operational time and higher costs, thereby limiting the practical applicability of electromagnetic imaging techniques \citep{2011Compressive}. In contrast, reducing the number of transmitters offers significant advantages, including lower costs and easier deployment in constrained environments.\citep{2007Compressive,2010Compressive,yang2015efficient}

However, the inherent ill-posed nature of \ac{eisp} poses significant challenges to accurate reconstruction~\citep{pan2011subspace,chen2018computational,li2019deepnis,zhong2016a, luo2024imaging, yao2019two, yao2020enhanced, yu2024pifon, wang2024universal}, particularly when only a limited number of transmitters are available. The scarcity of transmitters leads to an insufficient amount of measured data, which fails to capture adequate physical information for stable inversion. As a result, approaches relying solely on physical mechanisms\citep{slaney1984limitations, belkebir2005superresolution, chen2009subspace, zhong2011fft} often fail to achieve accurate reconstruction. Conventional numerical methods such as \ac{bp} \citep{belkebir2005superresolution}, generally fail to produce reliable reconstructions under such limited-data conditions. Recent machine learning-based approaches like \pg~\citep{song2021electromagnetic} and \pn~\citep{liu2022physics} often start with an initial solution derived from numerical methods, \ie, \ac{bp}, and frame the problem as an image-to-image translation task. With only a limited number of transmitters available, reliance on \ac{bp} becomes a critical bottleneck. When \ac{bp} fails, these methods are unable to correct its errors, as they are not fully end-to-end, ultimately leading to inaccurate reconstructions. The most recent method \iishort~\citep{luo2024imaging} integrates physical mechanisms into neural networks and performs case-by-case optimization. However, in limited-transmitter scenarios, even after optimization has converged, the resulting reconstructions may still diverge substantially from the ground truth (\cref{fig:revisit}), underscoring the intrinsic ambiguity of the inverse problem.

To address these limitations, we propose a \textit{fully end-to-end} and \textit{data-driven} framework that predicts the relative permittivity of scatterers from measured fields, leveraging data distribution priors to compensate for the scarcity of observational data. 
Unlike generative methods where data distribution priors typically refer to explicit, decoupled modules (e.g., denoisers modeling $p(x)$) \cite{kamilov2023, daras2024surveydiffusionmodelsinverse}, our method learns an end-to-end mapping that implicitly leverages data statistics to resolve the inherent ill-posedness of electromagnetic inverse scattering problems.
Specifically, our model takes the measured fields and the spatial coordinate of a position as input and directly predicts the relative permittivity at that location using \ac{mlp}s, and is trained in a fully end-to-end manner against the ground-truth data.
Our approach bypasses traditional numerical methods like \ac{bp}, thereby avoiding the inherent constraints associated with conventional inversion techniques in limited-transmitter scenarios and fully exploiting the advantages of data-driven learning.
This simple yet effective design enables efficient training across datasets and supports fast, feed-forward inference to achieve accurate and stable reconstruction predictions.

Extensive experiments demonstrate that our method outperforms existing \ac{sota} methods on multiple benchmark datasets, especially under the challenging single-transmitter setting, where all previous methods fail (\cref{fig:synthetic_0.05noise_N_inc=1}). It generalizes well to diverse scenarios and can be naturally extended to 3D scenes while maintaining high reconstruction accuracy. In summary, our contributions are threefold:

1) We systematically analyze the difficulty of lacking physical information faced by \ac{eisp} in the setting of few transmitters, and point out that the missing information can be supplemented by data distribution priors.

2) Based on our analysis, we propose a \textit{fully end-to-end} and \textit{data-driven} model that does not rely on traditional numerical methods.

3) Extensive experiments show that our method outperforms existing \ac{sota} approaches, especially under the challenging single-transmitter setting, marking a concrete step toward cost-effective and practical electromagnetic imaging solutions.

\vspace{-2mm}

\section{Related Work}
\label{sec:related}

\vspace{-1mm}

\subsection{Electromagnetic Inverse Scattering Problems}
Solving \ac{eisp} is to determine the relative permittivity of the scatterers based on the scattered field measured by the receivers, thereby obtaining internal imaging of the object. The primary challenges of \ac{eisp} arise from its nonlinearity, ill-posedness, and errors introduced by the discretization~\citep{pan2011subspace,chen2018computational,li2019deepnis,zhong2016a, luo2024imaging}. Traditional methods for solving \ac{eisp} can be categorized into non-iterative~\citep{slaney1984limitations, devaney1981inverse, habashy1993beyond, belkebir2005superresolution} and iterative~\citep{chen2009subspace, zhong2011fft, xu2017hybrid, habashy1994simultaneous, van1999extended} approaches. Non-iterative methods, such as the Born approximation~\citep{slaney1984limitations}, the Rytov approximation~\citep{devaney1981inverse, habashy1993beyond}, and the \ac{bp} method~\citep{belkebir2005superresolution}, solve nonlinear equations through linear approximations, which inevitably lead to poor quality of the results. 
For better reconstruction quality, iterative methods \citep{zhong2009twofold, chen2009subspace, zhong2011fft, xu2017hybrid, habashy1994simultaneous, van1999extended, gao2015sensitivity} such as 2-fold Subspace Optimization Method (SOM)~\citep{zhong2009twofold} and Gs SOM \citep{chen2009subspace} are proposed. To further overcome the ill-posedness of \ac{eisp}, diverse regularization approaches and prior information have been widely applied~\citep{oliveri2017compressive, shen2014sar, liu2018electromagnetic, anselmi2018iterative}. However, all of these methods are not generalizable and can be time-consuming because of the iterative schemes~\citep{liu2022physics}.  

\vspace{-1mm}

\begin{figure*}[t!]
    \vspace{-0.8cm}
    \centering
    \includegraphics[width=\linewidth]{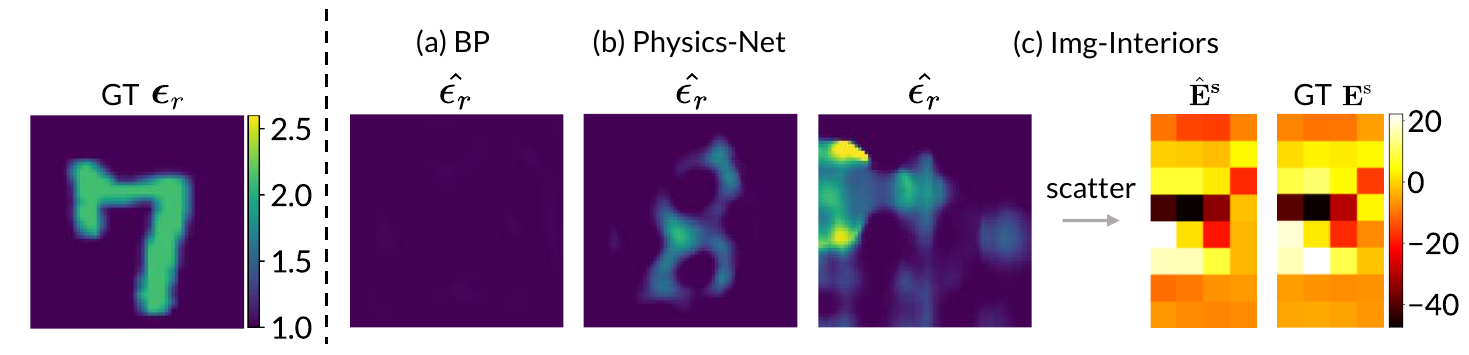}
    \caption{\textbf{Difficulties that previous methods faced under a single-transmitter setting.} (a) \ac{bp} cannot reconstruct the scatterer. (b) \pn makes incorrect guesses. (c) Although the reconstruction result of \iishort is consistent with the measured field, the reconstructed scatterer itself is completely different from the ground truth. }
    \label{fig:revisit}
     \vspace{-0.4cm}
\end{figure*}

\subsection{Machine Learning for \ac{eisp}}
Recent studies shift to leveraging neural networks to solve this problem and demonstrate promising results \citep{geng2021recent, li2024ifnet}.
Some work~\citep{wei2019deep, li2019deepnis, zhang2020learning, xu2021fast, sanghvi2019embedding, song2021electromagnetic, liu2022physics} adopt a two-stage strategy: they use non-iterative methods such as BP~\citep{belkebir2005superresolution} to generate initial estimates, which are then refined using image-to-image neural networks. While these approaches offer a degree of generalization, they are not end-to-end and remain dependent on BP initialization \citep{belkebir2005superresolution}, which becomes their bottleneck. When physical data are too insufficient to reconstruct the scatterer, especially under single-transmitter settings, these approaches tend to ``hallucinate'' outputs according to unreliable initialization rather than predict the scatterer based on measured field (see \cref{fig:synthetic_0.05noise_N_inc=1}). A more recent approach, \iishort~\citep{luo2024imaging}, integrates scattering mechanisms into the network architecture and achieves accurate reconstructions. However, it requires case-specific optimization, limiting generalization and making it vulnerable to local minima, often leading to failure in complex settings (see \cref{fig:synthetic_0.05noise_N_inc=16,fig:MNIST_0.30noise_N_inc=16}). Moreover, it fails under a single transmitter setting even when the optimization may have already converged because of ambiguity.
While our method is also learning-based, it is an end-to-end feed-forward framework that simultaneously achieves generalization through data-driven learning. As a result, it consistently outperforms \ac{sota} methods, particularly in the challenging single-transmitter setting where previous approaches fail.
\section{Revisiting \ac{eisp}}
\label{sec:revisit}
In this section, we revisit \ac{eisp} and uncover its fundamental challenge: the inherent ill-posedness stemming from information scarcity.

\vspace{4pt}
\noindent\textbf{Preliminary.} In the forward process, the transmitters produce incident electromagnetic field $\Eincg$ to the scatterer, generating scattered electromagnetic field $\Esg$. \ac{eisp} is the inverse problem of the forward process. That is, for an unknown scatterer, we use transmitters to apply certain incident field $\Eincg$ to it, and measure the scattered field $\Esg$ as our input via receivers. Our goal is to reconstruct the relative permittivity $\Epsrg$ throughout the scatterer. For a detailed introduction of \ac{eisp}, please refer to our supplementary material (\cref{supp:sec:eisp}).
Specifically, the incident field $\Eincg$ excites the induced current $\Jg$. Using the method of moments~\citep{peterson1998computational}, the total field $\Etotg$ for a given transmitter can be expressed as~\citep{colton2013integral}:
\begin{equation} \Etotg = \Eincg+\Gd \cdot \Jg, \label{state_equation} \end{equation}
where $\Etotg$ is a vector of length $M^2$, and $\Gd$ is a constant $M^2 \times M^2$ matrix representing the discrete free-space Green's function in $\D$. The induced current field $\Jg$ satisfies:
\begin{equation} \Jg = \operatorname{Diag}(\xig)\cdot\Etotg, \label{polarization} \end{equation}
where $\xig = \Epsrg - 1$, $\operatorname{Diag}(\xig)$ represents a diagonal matrix whose leading diagonal consists of $\xig$.
Then $\Jg$ serves as a new source to emit $\Esg$. For $\Nr$ receivers, the scattered field $\Esg$ can be got through $\Esg = \Gs\cdot\Jg$, where $\Gs$ is a constant $\Nr \times M^2$ matrix representing the discrete Green’s function. Since $\Nr \ll M^2$ in practice, reconstructing the induced current $\Jg$ from the scattered field $\Esg$ is ill-posed. 
\begin{figure*}[t!]
    \vspace{-0.8cm}
    \centering
    \includegraphics[width=\linewidth]{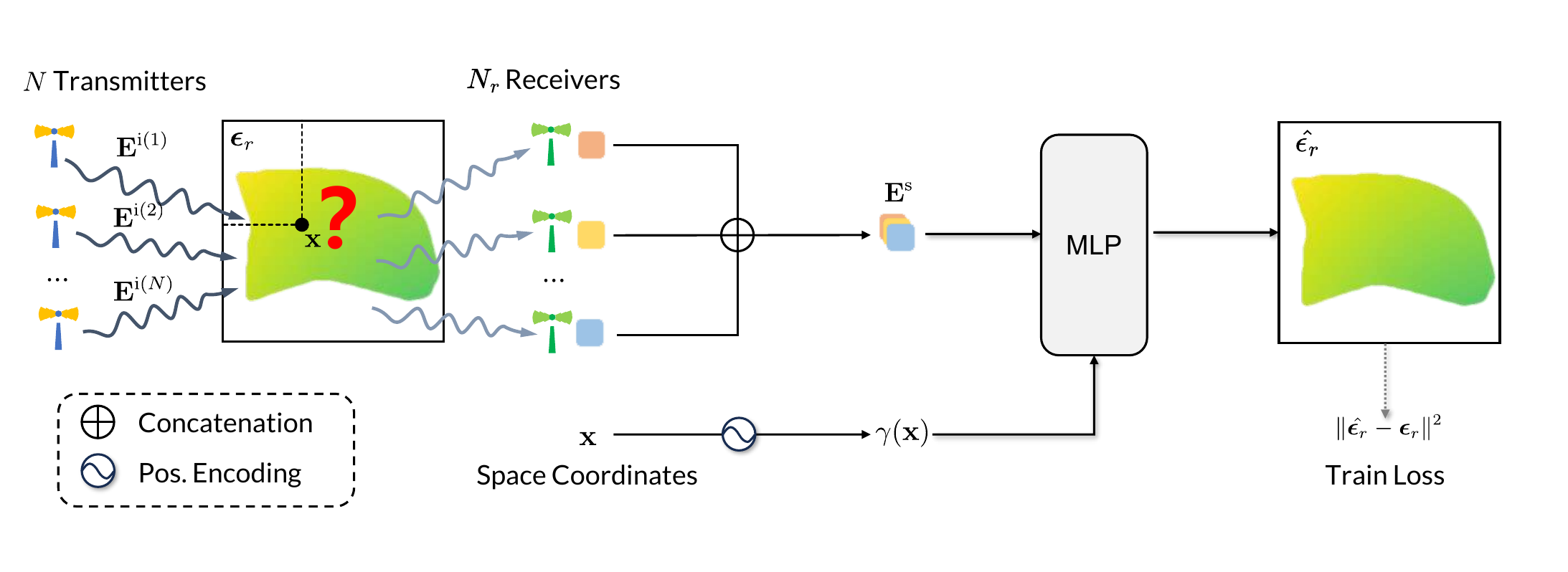}
    \vspace{-10mm}
    \caption{\textbf{Overview of our method.} Our pipeline is built around a \ac{mlp} that serves as the inverse solver. Given the scattered field measurements $\Esg$ from all transmitters and receivers, along with a spatial query $\x$, the \ac{mlp} directly predicts the relative permittivity $\hat{\Epsrg}(\x)$. To enhance spatial expressiveness, we apply positional encoding $\gamma(\x)$ to the query position. During training, dashed lines indicate the supervision signals applied. }
    \label{fig:pipeline}
\end{figure*}

\vspace{4pt}
\noindent\textbf{Reduction of measured data.} \ac{eisp} is fundamentally challenged by nonlinearity and ill-posedness, especially when the amount of measured data is significantly reduced, such as under single-transmitter settings. We divide previous work into three categories and systematically analyze the difficulties they faced under this setting. (a) Conventional numerical approaches, such as \ac{bp}~\citep{belkebir2005superresolution}, employ linear approximations, which limit their reconstruction quality. As shown in \cref{fig:revisit}, \ac{bp} cannot even reconstruct a rough shape of the scatterer. (b) Machine learning methods based on conventional numerical approaches \cite{song2021electromagnetic, liu2022physics}, such as \pn. Although \pn can leverage data-driven training to compensate for missing physical information, its strong dependency on \ac{bp} initialization becomes a critical bottleneck. When \ac{bp} fails, the model cannot correct the error of \ac{bp} because it is not fully end-to-end, resulting inaccurate reconstructions, as shown in \cref{fig:revisit}. (c) Machine learning methods based on implicit functions, such as \iishort~\citep{luo2024imaging}. \iishort reconstructs a scatterer through case-by-case optimization. As shown in \cref{fig:revisit}, we use the scatterer reconstructed by \iishort to simulate the scattered field, and the result closely matches the measured field. However, the scatterer itself deviates significantly from the ground truth, which shows the intrinsic ambiguity of the inverse problem. 
The core conclusion is that the severe information deficit makes a direct solution to the inverse problem fundamentally intractable. Consequently, any such attempt is bound to be fragile, highlighting the need for an alternative paradigm.

\section{Method}
\label{sec:method}

\subsection{Overview}
\label{sec:method:overview}

To address the aforementioned limitations, we introduce our end-to-end, data-driven framework for \ac{eisp}, as illustrated in \cref{fig:pipeline}.
Our method employs an \ac{mlp} that takes space coordinates $\x$ and corresponding scattered field measurements $\Esg$ as input, and directly outputs the relative permittivity $\Epsrg$ at the specified locations. This approach effectively learns the mapping between scattered field $\Esg$ and relative permittivity $\Epsrg$ through training on diverse scattering scenarios, thereby incorporating essential data distribution priors to compensate for the lack of physical information caused by insufficient measurements. In the following, we detail our model architecture (\cref{subsec:arch}), and the training losses (\cref{subsec:train}).
\subsection{Model Architecture}
\label{subsec:arch}
Based on the forward formulation of \ac{eisp} in \cref{sec:revisit}, where the scattered field measurements serve as input and the relative permittivity distribution represents the output, we design an end-to-end learning framework that directly learns this complex nonlinear mapping. As illustrated in \cref{fig:pipeline}, our approach employs an \ac{mlp} that approximates the inverse mapping from spatial coordinates and scattered field data to the relative permittivity values, formulated as:
\begin{equation}
    \hat{\Epsrg}(\xpoint) = \MLP (\Esg, \gamma(\xpoint)), \xpoint \in \mathbb{R}^2, 
\end{equation}
where $\xpoint$ represents the spatial coordinate, $\Esg$ denotes the scattered field measured by all receivers, $F_\theta(\cdot)$ is an \ac{mlp} with trainable parameters, and $\hat{\Epsrg}(\xpoint)$ is the predicted relative permittivity at the corresponding position. Recall that in \cref{sec:revisit}, for a single transmitter, the scattered field $\Esg$ is discretized as a real-valued vector of dimension $2\Nr$, containing the real and imaginary parts of the measurements from all $\Nr$ receivers. In the multiple transmitter configuration, $\Esg$ is constructed by combining the complex measurement data from all $\Ni$ transmitters, resulting in a real-valued vector of dimension $2\Ni \cdot \Nr$ that represents the wave propagation and scattering behavior under diverse illumination conditions provided by transmitters at different locations.
To enhance the model's capacity to represent high-frequency features, we apply positional encoding to the spatial coordinates $\xpoint$, mapping them into a higher-dimensional Fourier feature space using the encoding function: $\gamma(x) = [\sin(x), \cos(x), \ldots, \sin(2^{\Omega-1}x), \cos(2^{\Omega-1}x)]$, where the hyperparameter $\Omega$ controls the spectral bandwidth.
The complete relative permittivity distribution $\hat{\Epsrg}$ is reconstructed by sampling the MLP at all grid points $\{\xpoint\}_{i=1}^{M^2}$: $\hat{\Epsrg} = \{ \MLP (\Esg, \gamma(\xpoint))\}_{\text{i}=1}^{M^2}.$
\subsection{Training}
\label{subsec:train}
Our training objective is defined by a single loss function designed to directly supervise the reconstruction accuracy of the relative permittivity distribution. The loss is formulated as: $\Lall = \|\hat{\Epsrg} - \Epsrg\|^2.$
where $\hat{\Epsrg}$ denotes the predicted relative permittivity and $\Epsrg$ represents the ground truth. By minimizing this \ac{mse} loss between the predicted and true permittivity values, the model learns to infer the material properties directly from the scattered field measurements, effectively leveraging the data distribution priors to overcome the ill-posedness of the inverse problem. This simplified loss function ensures stable and efficient training.
\begin{table*}[t]
\vspace{-7pt}
 \centering
\caption{\textbf{Quantitative comparison results with \ac{sota} methods.} For \ac{circular} and \ac{mnist} datasets, we report results under two noise levels: 5\% and 30\%. The best results are shown in \textbf{bold}, and the second-best results are \underline{underlined}.}
\vspace{-1mm}
\label{tab:result}
\resizebox{\linewidth}{!}{
\setlength{\tabcolsep}{0.2mm}
\begin{tabular}{l|ccc|ccc|ccc|ccc|ccc}
\thickhline
\multirow{2}{*}{Method} & \multicolumn{3}{c|}{MNIST (5\%)} & \multicolumn{3}{c|}{MNIST (30\%)} & \multicolumn{3}{c|}{Circular (5\%)} & \multicolumn{3}{c|}{Circular (30\%)}  & \multicolumn{3}{c}{IF} \\
& MSE $\downarrow$         & SSIM $\uparrow$        & PSNR $\uparrow$        & MSE $\downarrow$        & SSIM $\uparrow$         & PSNR $\uparrow$        & MSE $\downarrow$       & SSIM $\uparrow$       & PSNR $\uparrow$       & MSE $\downarrow$       & SSIM $\uparrow$        & PSNR $\uparrow$  & MSE $\downarrow$       & SSIM $\uparrow$        & PSNR $\uparrow$ \\  
\thickhline
\rowcolor{mygray}
\multicolumn{13}{c|}{Number of Transmitters: $N=16$} & \multicolumn{3}{c}{$N=8/18$} \\
BP  \cite{belkebir2005superresolution}    & 0.177        & 0.719       & 16.43       & 0.178        & 0.716        & 16.38     & 0.052         & 0.905        & 27.41        & 0.053         & 0.904         & 27.42      & 0.190 & 0.779 & 16.19       \\
\tfshort \cite{zhong2009twofold}     & 0.154        & 0.757       & 20.93       & 0.156        & 0.738        & 20.84   & 0.031         & 0.917        & 32.23        & 0.038         & 0.889         & 30.63     & \text{-}  & \text{-}  & \text{-} \\
Gs SOM \cite{chen2009subspace}    & \underline{0.072}        & 0.923       & \underline{28.31}       & \underline{0.081}        & 0.901        & \underline{27.13}  & \underline{0.023}         & 0.946        & 35.40        & \textbf{0.024}         & 0.937         & \underline{34.89}      & 0.184 & 0.790 & 17.00    \\
BPS  \cite{chen2018computational, wei2019deep}     & 0.093        & 0.909       & 25.00       & 0.105        & 0.891        & 23.90     & 0.027         & 0.963        & 33.00        & \underline{0.029}         & \textbf{0.956}         & 32.42      & 0.310 & 0.664 & 17.05  \\
PGAN \cite{song2021electromagnetic}           & 0.084        & 0.916       & 25.80       & 0.091        & \underline{0.910}        & 25.31     & 0.026         & \textbf{0.966}        & \underline{35.56}        & 0.032         & 0.947         & 33.91 & \underline{0.121} & \textbf{0.926} & \underline{24.78}        \\
Physics-Net \cite{liu2022physics}  & 0.075        & \underline{0.932}       & 26.17       & 0.093        & 0.906        & 24.58 & 0.027         & 0.934        & 32.72        & 0.030         & 0.927         & 32.08        & 0.170 & 0.788 & 18.48     \\
Two-Step \cite{yao2019two}  &  0.111       &  0.835      &  23.17      & 0.111        &  0.835       & 22.91 & 0.059        &  0.830      &  25.92       &    0.091      &    0.848       &  23.94       &\text{-}  & \text{-} &  \text{-}    \\
\iishort \cite{luo2024imaging}     & 0.200        & 0.863       & 26.41       & 0.336        & 0.760        & 19.01  & 0.036         & 0.947        & 35.05        & 0.047         & 0.932         & 32.62        & 0.153 & 0.837 & 23.26    \\    
 \textbf{Ours}   &         \textbf{0.039}&        \textbf{0.978}&        \textbf{32.11}&         \textbf{0.050}&         \textbf{0.966}&        \textbf{29.91}  &  \textbf{0.020}&        \underline{0.965}  &         \textbf{36.92}&           \textbf{0.024}&         \underline{0.954} &         \textbf{35.19}& \textbf{0.094} & \underline{0.916} & \textbf{24.89} \\
\thickhline
\rowcolor{mygray}
\multicolumn{16}{c}{Number of Transmitters: $N=1$} \\
BP \cite{belkebir2005superresolution} & 0.194& 0.698& 15.40& 0.194& 0.696& 15.40 & 0.065& 0.892& 25.30& 0.065& \underline{0.892}& 25.30 & 0.199& 0.770& 16.29\\
\tfshort \cite{zhong2009twofold}& 0.432& 0.556& 12.49& 0.828& 0.382& 9.45 & 0.060& 0.859& 26.63& 0.157& 0.639& 20.07 & \text{-} &\text{-} &\text{-} \\
Gs SOM \cite{chen2009subspace}& 0.460& 0.598& 15.31& 0.404& 0.557& 14.91 & 0.046& 0.888& 29.62& 0.051& 0.862& 28.77 & 0.192& \underline{0.779}& 16.66\\
BPS \cite{chen2018computational, wei2019deep}& 0.189& 0.774& 18.75& 0.205& 0.744& 17.97 & 0.045& 0.891& 29.29& 0.055& 0.862& 27.68 & 0.348& 0.669& 16.18\\
PGAN \cite{song2021electromagnetic}& 0.133& \underline{0.867}& 21.69& 0.153& \underline{0.830}& \underline{20.41} & \underline{0.033}& \textbf{0.932}& \underline{32.02}& \underline{0.040}& \textbf{0.914}& \underline{29.94}& 0.248& 0.680& 16.85\\
PhysicsNet \cite{liu2022physics}& 0.137& 0.798& 19.98& \underline{0.152}& 0.783& 19.38 & 0.055& 0.887& 26.60& 0.056& 0.890& 26.48 & \underline{0.175}& 0.771& \underline{17.45}\\
Two-Step \cite{yao2019two}& \underline{0.117}& 0.845& \underline{23.34}& 0.203& 0.656& 17.62 & 0.145& 0.673& 19.45& 0.174& 0.652& 18.41 & \text{-}& \text{-}& \text{-}\\
\iishort \cite{luo2024imaging} & 0.305& 0.604& 16.06& 0.467& 0.484& 12.47& 0.096& 0.855& 26.19& 0.153& 0.806& 20.90& 0.305& 0.705& 17.34\\
\textbf{Ours} & \textbf{0.085}& \textbf{0.921}& \textbf{26.09}& \textbf{0.127}& \textbf{0.862}& \textbf{22.56}& \textbf{0.031} & \underline{0.931}& \textbf{33.18}& \textbf{0.038} & \textbf{0.914}& \textbf{31.38}& \textbf{0.128}& \textbf{0.908}& \textbf{24.19}\\
    \thickhline
 \end{tabular}}
 \vspace{-0.2cm}
\end{table*}
\vspace{-2mm}

\section{Experiments}
\label{sec:exp}
\subsection{Setup}
\noindent\textbf{Datasets.}
We train and test our method on standard benchmarks used for \ac{eisp} following previous work \citep{wei2019deep, song2021electromagnetic, liu2022physics}. Datasets that share identical transmitter and receiver configurations are combined into a unified training set. 1) Synthetic \ac{circular}~\citep{luo2024imaging} is synthetically generated comprising images of cylinders with random relative radius, number, location, and permittivity. 2) Synthetic \ac{mnist}~\citep{deng2012mnist} contains grayscale images of handwritten digits. For the two synthetic datasets, following previous work\citep{luo2024imaging, deng2012mnist}, we evaluate two levels of noise: 5\% and 30\%, and the number of receivers $\Nr=32$.
3) Real-world \ac{ifdata}~\citep{2005InvPr} contains three different dielectric scenarios, namely \fde, \fdi, and \ftd, where $\Nr=241$. 4) Synthetic \ac{3dmnist}~\citep{3Dmnist} contains 3D data of handwritten digits. 5) \ac{3dshapenet}~\citep{wu20153d} contains 3D data of various shapes. $\Nr=160$ for the two 3D datasets. For more details about datasets, please refer to \cref{supp:sec:setup}.

\vspace{2pt}
\noindent\textbf{Baselines and Metrics.} 
For a fair comparison, we follow the same setting as in previous work~\citep{wei2019deep,song2021electromagnetic,sanghvi2019embedding} using their official code and train or optimize each method under the same setup as ours \footnote{BPS, Physics-Net, PGAN and Two-Step are trained as they are learning-based methods; all other baselines are optimization-based; we train separate models for each noise level for both our method and other learning-based models. }. Specifically, we compare our method with 3 traditional methods and 4 deep learning-based approaches: 1) \textbf{BP}~\citep{belkebir2005superresolution}: A traditional non-iterative inversion algorithm. 2) \textbf{\tfshort}~\citep{zhong2009twofold}: A traditional iterative minimization scheme by using SVD decomposition. 3) \textbf{Gs SOM}~\citep{chen2009subspace}: A traditional subspace-based optimization method by decomposing the operator of Green's function. 4) \textbf{BPS}~\citep{chen2018computational, wei2019deep}: A CNN-based image translation method with an initial guess from the BP algorithm. 5) \textbf{Physics-Net}~\citep{liu2022physics}: A CNN-based approach that incorporates physical phenomena during training. 6) \textbf{Two-Step}~\citep{yao2019two}: A CNN-based approach with two steps. 7) \textbf{PGAN}~\citep{song2021electromagnetic}: A CNN-based approach using a generative adversarial network. 8) \textbf{\iishort}~\citep{luo2024imaging}: An implicit approach optimized by forward calculation. 
Following previous work~\citep{liu2022physics}, we evaluate the quantitative performance of our method using PSNR \citep{wang2009mean}, SSIM \citep{wang2004image}, and \ac{rrmse}~\citep{song2021electromagnetic}. 
\subsection{Comparison with \ac{sota}s}
\subsubsection{Multiple Transmitter Evaluation}
\begin{figure*}
    \centering
    \includegraphics[width=1\linewidth]{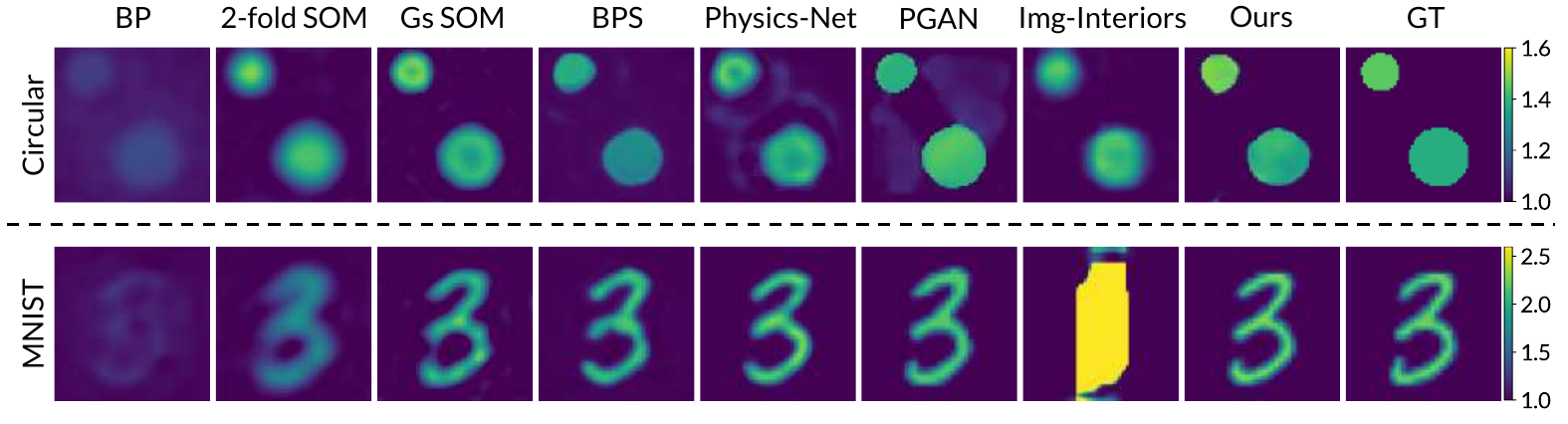}
    \vspace{-0.8cm}
    \caption{\textbf{Qualitative comparison under the multiple-transmitter setting.} The results are obtained with $\Ni=16$ transmitters and a noise level of 5\%. Colors represent the values of the relative permittivity.}
    \label{fig:synthetic_0.05noise_N_inc=16}
\end{figure*}
\begin{figure*}[h]
    \centering
    \includegraphics[width=\linewidth]
    {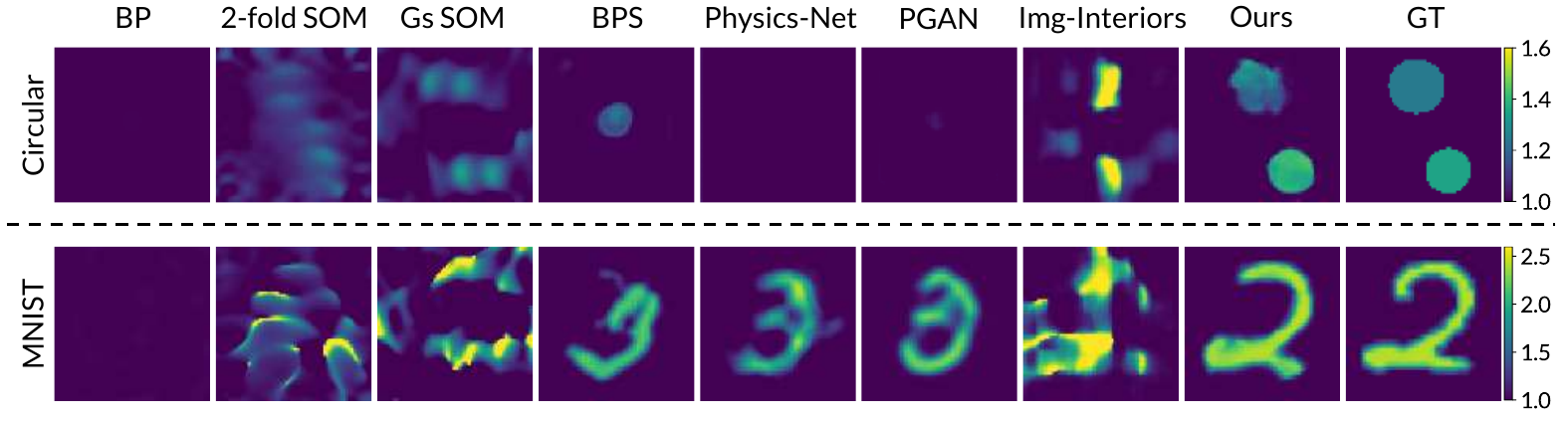}
    \vspace{-0.8cm}
    \caption{\textbf{Qualitative comparison under the single-transmitter setting.} Results are obtained with $\Ni=1$ transmitter and noise level of 5\%. Colors represent the values of the relative permittivity.}
    \label{fig:synthetic_0.05noise_N_inc=1}
\end{figure*}
We begin by comparing our method against prior approaches under the multiple-transmitter setting, using both synthetic and real datasets for comprehensive evaluation.
As shown in the upper part of \cref{tab:result}, our method achieves comparable or superior performance to the \ac{sota} in most cases, demonstrating how our end-to-end training framework successfully leverage the data prior across diverse data domains.
In addition, we present a qualitative comparison, as shown in \cref{fig:synthetic_0.05noise_N_inc=16}. Traditional methods such as \ac{bp}\citep{belkebir2005superresolution}, \gs\citep{chen2009subspace}, and \tfshort~\citep{chen2009subspace} are only capable of recovering the coarse shape of the scatterer. \bps~\citep{chen2018computational, wei2019deep} produces sharp edges, but the reconstructed shapes are often inaccurate. \pg~\citep{song2021electromagnetic} achieves accurate shape recovery, yet introduces noticeable background artifacts. \iishort~\citep{luo2024imaging} can generate high-quality reconstructions, but occasionally fails due to local optima, as it is based on an iterative optimization process (see the last row). In contrast, our method produces accurate and clean reconstructions across all cases, demonstrating both visual fidelity and robustness.
\vspace{-1mm}
\subsubsection{Single Transmitter Evaluation}
\begin{figure*}
\vspace{-0.2cm}
    \centering
    \includegraphics[width=.97\linewidth]{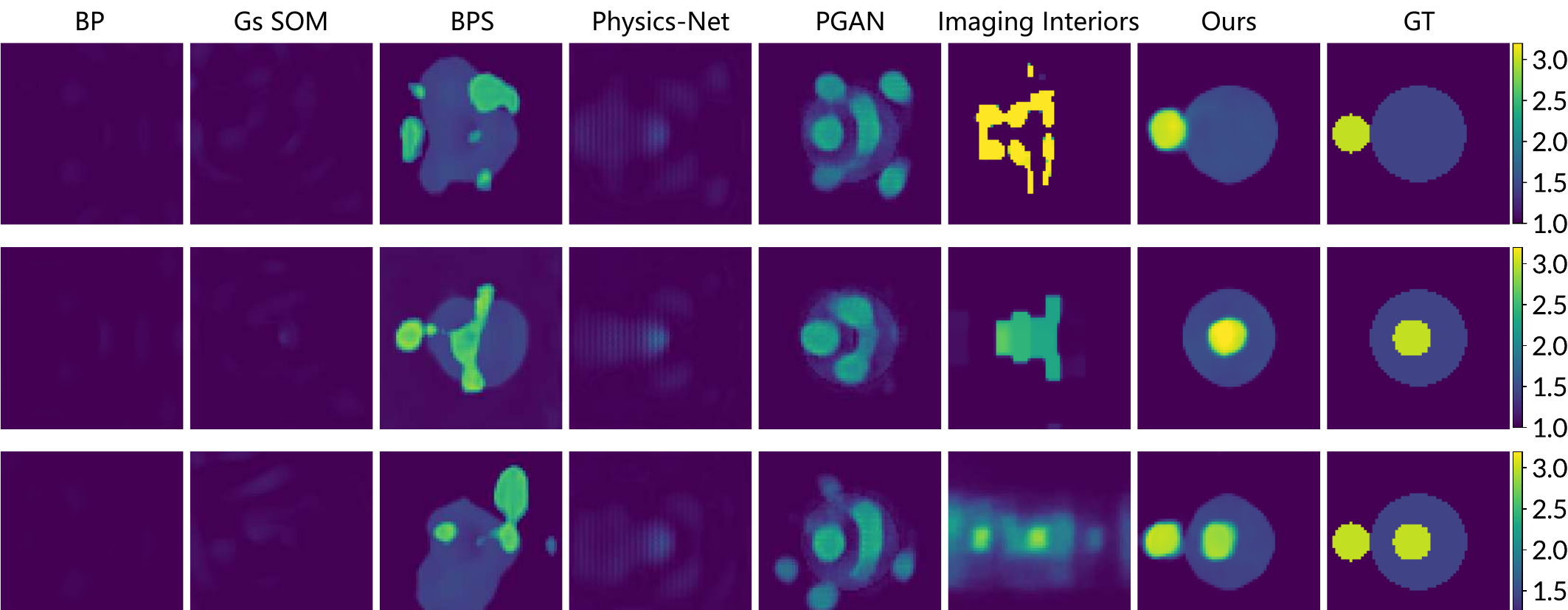}
    \caption{\textbf{Qualitative comparison under single-transmitter setting on real-world \ac{ifdata} dataset.} The results are obtained with $\Ni=1$ transmitter, without noise level. Colors represent the values of the relative permittivity.}
    \label{fig:IF_N_inc=1}
\end{figure*}
Furthermore, we investigate a highly challenging and practically important setting that has been largely underexplored in previous work: performing EISP with a minimal number of transmitters. Specifically, we consider the most extreme case, using only a single transmitter. As shown in the lower part of \cref{tab:result}, our method significantly outperforms all previous approaches across all datasets and noise levels. This remarkable performance under such constrained conditions underscores the efficacy of our end-to-end training framework, which successfully encodes and leverages rich data priors to achieve state-of-the-art results across diverse domains.
To better understand this phenomenon, we present qualitative comparisons in \cref{fig:synthetic_0.05noise_N_inc=1} for synthetic data and \cref{fig:IF_N_inc=1} for real-world \ac{ifdata}~\citep{2005InvPr}. Traditional methods such as \ac{bp}\citep{belkebir2005superresolution}, \gs\citep{chen2009subspace}, and \tfshort~\citep{chen2009subspace} produce only blurry reconstructions. Deep learning-based methods like \bps~\citep{chen2018computational, wei2019deep}, \pn~\citep{liu2022physics}, and \pg~\citep{song2021electromagnetic} tend to “hallucinate” the digit, resulting in wrong shape on the \ac{mnist} dataset. \iishort~\citep{luo2024imaging} fails to capture the fundamental morphology of the scatterer, resulting in structurally inaccurate representations that deviate significantly from the ground truth. Among all the methods, only ours can still produce reasonably accurate reconstructions of the relative permittivity under such an extreme condition.

\begin{figure*}[h]
    \centering
    \includegraphics[width=\linewidth]{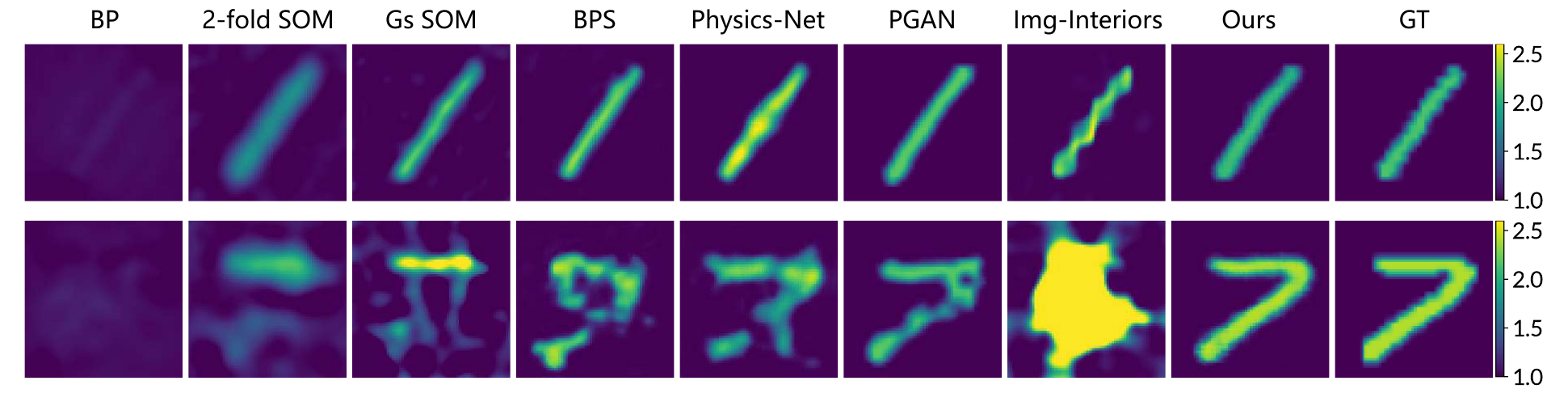}
    \vspace{-0.8cm}
\caption{\textbf{Qualitative comparison under high noise setting.} The results are obtained with $\Ni=16$ transmitters and a noise level of 30\%. Colors represent the values of the relative permittivity.}
\label{fig:MNIST_0.30noise_N_inc=16}
\vspace{-0.3cm}
\end{figure*}
\subsection{Ablation Study}
\noindent\textbf{Noise Robustness.} To simulate real-world sensor noise and related perturbations, we evaluate the robustness of the models by adding noise to the scattered field. Moving beyond simple binary testing, we systematically assess the model's performance across multiple noise levels ranging from 5\% to 30\% to examine its behavior in various noisy environments. The quantitative results presented in \cref{tab:noise_ablation} demonstrate that our model exhibits smooth and gradual performance degradation as the noise level increases, maintaining excellent reconstruction capability even under strong noise interference as high as 30\%. Qualitative visualizations in \cref{fig:MNIST_0.30noise_N_inc=16} show that most baseline methods exhibit noticeable artifacts or even complete failure under severe noise conditions, while our method remains robust and preserves the essential structure of the target. 
\begin{table*}[t!]
\centering
\begin{minipage}[t]{0.42\textwidth}
\centering
\caption{\textbf{Ablation study of noise levels effects on \ac{mnist} under the multiple-transmitter setting.}}
\label{tab:noise_ablation}
\begin{footnotesize}
\resizebox{\linewidth}{!}{
\setlength{\tabcolsep}{3.2mm}
\begin{tabular}{l |cccc|c}
\thickhline
Noise Level & \ac{rrmse} $\downarrow$ & SSIM$\uparrow$ & PSNR$\uparrow$ \\
\hline 
5\%  & 0.039 & 0.978 & 32.11 \\ 
10\% & 0.039 & 0.978 & 32.18 \\
15\% & 0.043 & 0.973 & 31.30 \\
20\% & 0.043 & 0.974 & 31.34 \\
25\% & 0.046 & 0.970 & 30.59 \\
30\% & 0.050 & 0.966 & 29.91 \\
\thickhline
\end{tabular}}
\end{footnotesize}
\vspace{-4pt}
\label{tab:ablation_noise}
\end{minipage}
\hfill 
\begin{minipage}[t]{0.54\textwidth}
\centering
\caption{\textbf{Ablation study on training data size under the mutiple-transmitter setting.} Noise levels ($5\%$ and $30\%$) in parentheses.}
\label{tab:data_size_ablation}
\begin{footnotesize}
\resizebox{\linewidth}{!}{
\setlength{\tabcolsep}{2.0mm}
\begin{tabular}{l |ccc|ccc}
\thickhline
\multirow{2}{*}{Data} & \multicolumn{3}{c|}{MNIST (5\%)} & \multicolumn{3}{c}{MNIST (30\%)} \\ 
& \ac{rrmse} $\downarrow$ & SSIM$\uparrow$ & PSNR$\uparrow$ & \ac{rrmse} $\downarrow$ & SSIM$\uparrow$ & PSNR$\uparrow$ \\
\hline 
100\% & 0.039 & 0.978 & 32.11 & 0.050 & 0.966 & 29.91 \\
75\% & 0.043 & 0.974 & 31.63 & 0.059 & 0.956 & 28.77 \\
50\% & 0.048 & 0.968 & 30.68 & 0.068 & 0.944 & 27.69 \\
25\% & 0.064 & 0.948 & 28.89 & 0.101 & 0.902 & 25.44 \\
\thickhline
\end{tabular}}
\end{footnotesize}
\vspace{-4pt}
\label{tab:ablation_datasize}
\end{minipage}
\end{table*}

\vspace{4pt}
\noindent\textbf{Ablation on Training Data Size.} To investigate the dependency of model performance on training data volume, we trained our model on varying scales of data from 100\% down to 25\% and evaluated them on a complete test set. The quantitative results are presented in \cref{tab:ablation_datasize}.
First, our model demonstrates remarkable data efficiency, maintaining strong performance even when trained on partial datasets. Second, the performance degradation becomes substantially more pronounced under high-noise conditions. The performance penalty for data reduction is markedly severer in high-noise scenarios. This pronounced contrast underscores that sufficient training data is crucial for the model to learn robust features capable of countering strong noise interference.
\subsection{Reconstruction on 3D data}
\begin{figure*}[t!]
\vspace{0.1cm}
    \centering
    \includegraphics[width=\linewidth]{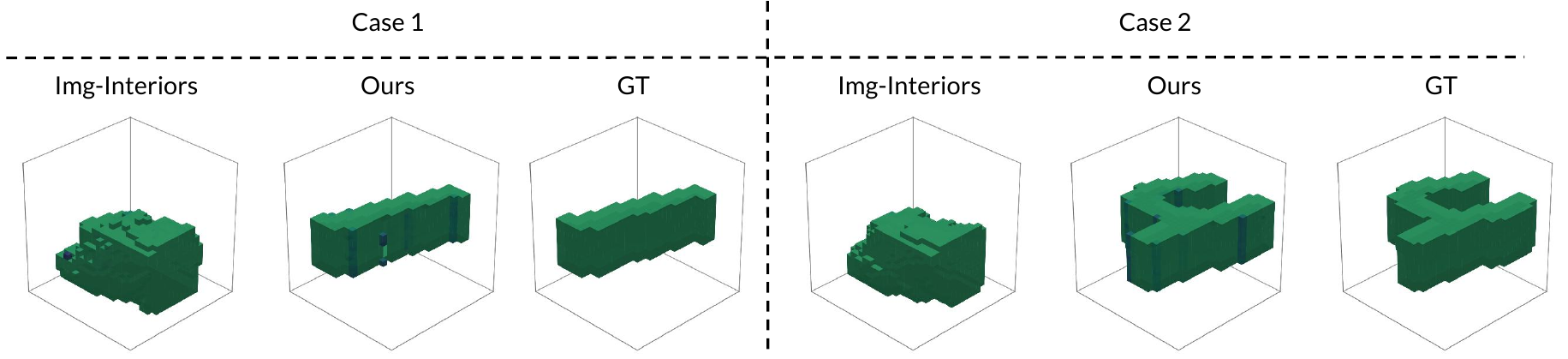}
\caption{\textbf{Qualitative comparison under the single-transmitter setting for 3D reconstruction on \ac{3dmnist} dataset.} The results are obtained with a single transmitter ($\Ni=1$). The voxel colors represent the values of the relative permittivity.}
    \label{fig:3D_fid}
\end{figure*}

\begin{figure*}[t!]
    \centering
    \includegraphics[width=\linewidth]{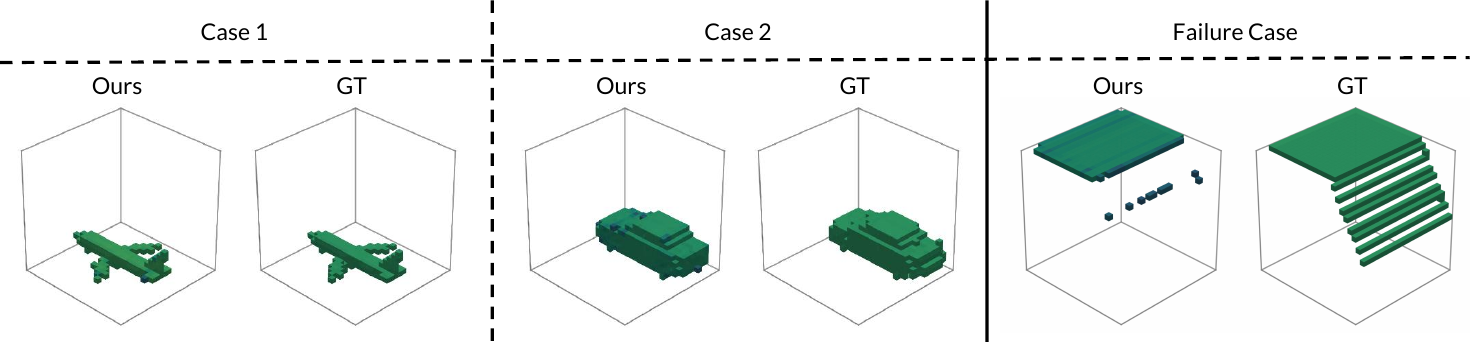}
\caption{\textbf{Qualitative comparison under the single-transmitter setting for 3D reconstruction on \ac{3dshapenet} dataset.} The results are obtained with a single transmitter ($\Ni=1$). The voxel colors represent the values of the relative permittivity.}
    \label{fig:3D_fid_ShapeNet}
    \vspace{-0.2cm}
\end{figure*}

\noindent\textbf{Setup and Metrics.}
Our method can be naturally extended to 3D scenarios. We use the same \ac{mlp} architecture, with the input dimension extended to 3. Following previous work \citep{luo2024imaging}, we employ the Synthetic \ac{3dmnist}~\citep{3Dmnist} and extend to \ac{3dshapenet}~\citep{wu20153d} for training and testing. For evaluation, we adopt 3D versions of the \ac{rrmse}~\citep{song2021electromagnetic} and \ac{iou} as our metrics. Further details on the datasets are provided in \cref{supp:sec:setup}.

\noindent\textbf{Results.} We evaluate our method and \iishort under limited-transmitter settings. Quantitative results demonstrate the superiority of our approach: on \ac{3dmnist}, our method achieves an \ac{mse} of 0.120 and \ac{iou} of 0.769 with $\Ni=1$ transmitter, significantly outperforming \iishort which obtains an \ac{mse} of 0.372 and \ac{iou} of 0.094 under the same conditions. With $\Ni=6$ transmitters, our results further improve to \ac{mse} of 0.094 and \ac{iou} of 0.834. For the more complex \ac{3dshapenet} dataset under $\Ni=1$ configuration, our method obtains an \ac{mse} of 0.064 and \ac{iou} of 0.762, showcasing its generalization capability to diverse 3D structures. \cref{fig:teaser} provides a comprehensive comparison of reconstruction quality and runtime between the two methods for both $\Ni=1$ and $\Ni=6$ configurations. \cref{fig:3D_fid} and \cref{fig:3D_fid_ShapeNet} provide visual comparisons of additional 3D reconstruction results on the 3D MNIST and 3D ShapeNet datasets. These results show that our method maintains robustness and generalizes effectively across geometrically complex 3D structures, representing significant progress towards practical applications.

\section{Conclusion}
\label{sec:conclusion}

In this work, we propose a fully end-to-end data-driven framework for electromagnetic inverse scattering that directly predicts relative permittivity from scattered field measurements. By leveraging data distribution priors to compensate for the lack of physical information, our method demonstrates state-of-the-art reconstruction accuracy and robustness, particularly in challenging single-transmitter scenarios where existing methods fail. This work highlights the potential of data-driven approaches to overcome the ill-posedness of inverse problems and provides a practical path toward cost-effective electromagnetic imaging.

\vspace{4pt}
\noindent\textbf{Limitations.} While our method effectively handles sparse transmitter settings, it struggles to reconstruct thin structures (see the rightmost block in \cref{fig:3D_fid_ShapeNet} for a typical failure case) and cannot accommodate varying receiver or transmitter locations. These limitations challenge deployment in real-world scenarios, where sensor layouts and environmental complexity vary. Addressing fine-structure reconstruction and flexible sensor configurations remains an important direction for future work.

\clearpage
\section*{Acknowledgments}
This work was supported by NSFC-6247070125, Ningbo Grants 2025Z038 and 2025Z059, and the High Performance Computing Center at Eastern Institute of Technology and Ningbo Institute of Digital Twin.

{
    \small
    \bibliographystyle{ieeenat_fullname}
    \bibliography{main}
}
\appendix
\clearpage

This supplementary material provides additional details and results, including the use of LLMs (\cref{supp:sec:llm}), physical model of Electromagnetic Inverse Scattering Problems (EISP) (\cref{supp:sec:eisp}), the experimental setup (\cref{supp:sec:setup}), and extended experimental results (\cref{supp:sec:exp}).

\section{The Use of Large Language Models (LLMs)} \label{supp:sec:llm}
We use GPT-4 solely for the purpose of polishing our language of manuscript. This includes improving grammatical accuracy and sentence fluency. LLMs play no significant role in research ideation, methodology design, and experiment execution of this paper. 

\section{Physical Model of EISP} \label{supp:sec:eisp}
The research subject of \ac{eisp} is scatterers. We can describe scatterers with their relative permittivity $\Epsr(\x)$. The relative permittivity is a physical quantity determined by the material, and it represents the ability to interact with the electromagnetic field of the material. The relative permittivity of vacuum is $1$ (and the relative permittivity of air is almost $1$), while the relative permittivity of scatterers is over $1$. The relative permittivity of metal material is positive infinity, so metal can shield against electromagnetic waves. Thus, scatterers are not composed of metal. 

In \ac{eisp}, the scatterer is placed within the region of interest $\D$. The transmitters are placed around $\D$, emitting incident electromagnetic field $\Einc$. The incident field then interacts with the scatterer, exciting the induced current $\J$. The induced current can act as secondary radiation sources, emitting scattered field $\Es$. In fact, for a point in the scatterer, it cannot distinguish the incident field and the scattered field at this point. So, it interacts with the sum of the incident field and the scattered field, aka the total field $\Etot$. The total field can be described by Lippmann-Schwinger equation \citep{colton2013integral} as follows: 
\begin{equation}
    \label{state_equation_conti}
    \Etot(\x)=\Einc(\x)+\kz^2\int_\D g(\x,\xp)\J(\xp)d\xp,\x\in\D,
\end{equation}
where $\x$ and $\xp$ are the spatial coordinates. $\kz$ is the wavelength of the electromagnetic wave determined by the frequency. $g(\x,\xp)$ is the free space Green's function, which represents the impact of the induced current $\J$ at the point $\xp$ to the total field at the point $\x$. $\x\in\D$ indicates that the total field is with the region of interest $\D$. The relationship between the induced current $\J$ and the total field $\Etot$ can be expressed as follows: 
\begin{equation}
    \label{polarization_conti}
    \J(\x)=\xi(\x)\Etot(\x),
\end{equation}
where $\xi(\x)=\Epsr(\x)-1$. 

The induced current generates scattered field, and we can measure the scattered field with receivers around the region of interest $\D$. The scattered field can be expressed as follows: 
\begin{equation}
    \label{data_equation_conti}
    \Es=\kz^2\int_\D g(\x,\xp)\J(\xp)d\xp,\x\in S,
\end{equation}
where $\x\in S$ indicates that the scattered field is measured by the receivers at surface $S$ around $\D$.  

Since digital analysis only applies to discrete variables~\citep{mitra2001digital, taflove2005computational}, we discretize equations \cref{state_equation_conti}, \cref{polarization_conti}, and \cref{data_equation_conti}. The region of interest $\D$ is discretized into $M\times M$ square subunits, and we use the method of moment \citep{peterson1998computational} to obtain the discrete scattered field $\Esg$\footnote{We use the \textbf{bold letters} to represent discrete variables.}. The discrete version of \cref{state_equation_conti} is as follows: 
\begin{equation} \Etotg = \Eincg+\Gd \cdot \Jg, \label{state_equation_supp} \end{equation}
where $\Gd$ is the discrete free space Green's function from points in the region of interest to points in the region of interest, which is a matrix of the shape $M^2\times M^2$. The discrete version of \cref{polarization_conti} is as follows: 
\begin{equation} \Jg = \operatorname{Diag}(\xig)\cdot\Etotg. \label{polarization_supp} \end{equation}
And the discrete version of \cref{data_equation_conti} is as follows:
\begin{equation}\Esg = \Gs\cdot\Jg, \label{data_equation_supp}\end{equation}
where $\Gs$ is the discrete free space Green's function from points in the region of interest to the locations of receivers, which is a matrix of the shape $\Nr\times M^2$. 
\section{Details of Experimental Setup} \label{supp:sec:setup}

\subsection{Datasets}
The datasets utilized in this work are described in detail below. To enhance model robustness and training efficiency, datasets that share identical measurement settings are pooled to form consolidated training sets. We strictly follow the established rules in \ac{eisp}~\citep{wei2019deep, song2021electromagnetic, liu2022physics} to set the dataset. We train and test our model on standard benchmarks used for \ac{eisp}. 

1) \ac{circular} \citep{luo2024imaging} is synthetically generated comprising images of cylinders with random relative radius, number, location, and permittivity between $1$ and $1.5$. $10k$ images are generated for training purposes, and $1.2k$ images for testing. 

2) \ac{mnist} \citep{deng2012mnist} contains grayscale images of handwritten digits. Similar to previous settings~\citep{wei2019deep, zhou2022deep}, we use them to synthesize scatterers with relative permittivity values between $2$ and $2.5$ according to their corresponding pixel values. The entire MNIST training set containing $60k$ images is used for training purposes, while $1.2k$ images from the MNIST test set are randomly selected for testing. 

\begin{figure*}
    \centering
    \subfloat[\label{fig:schematic_synthetic}]{
        \includegraphics[width=0.49\linewidth]{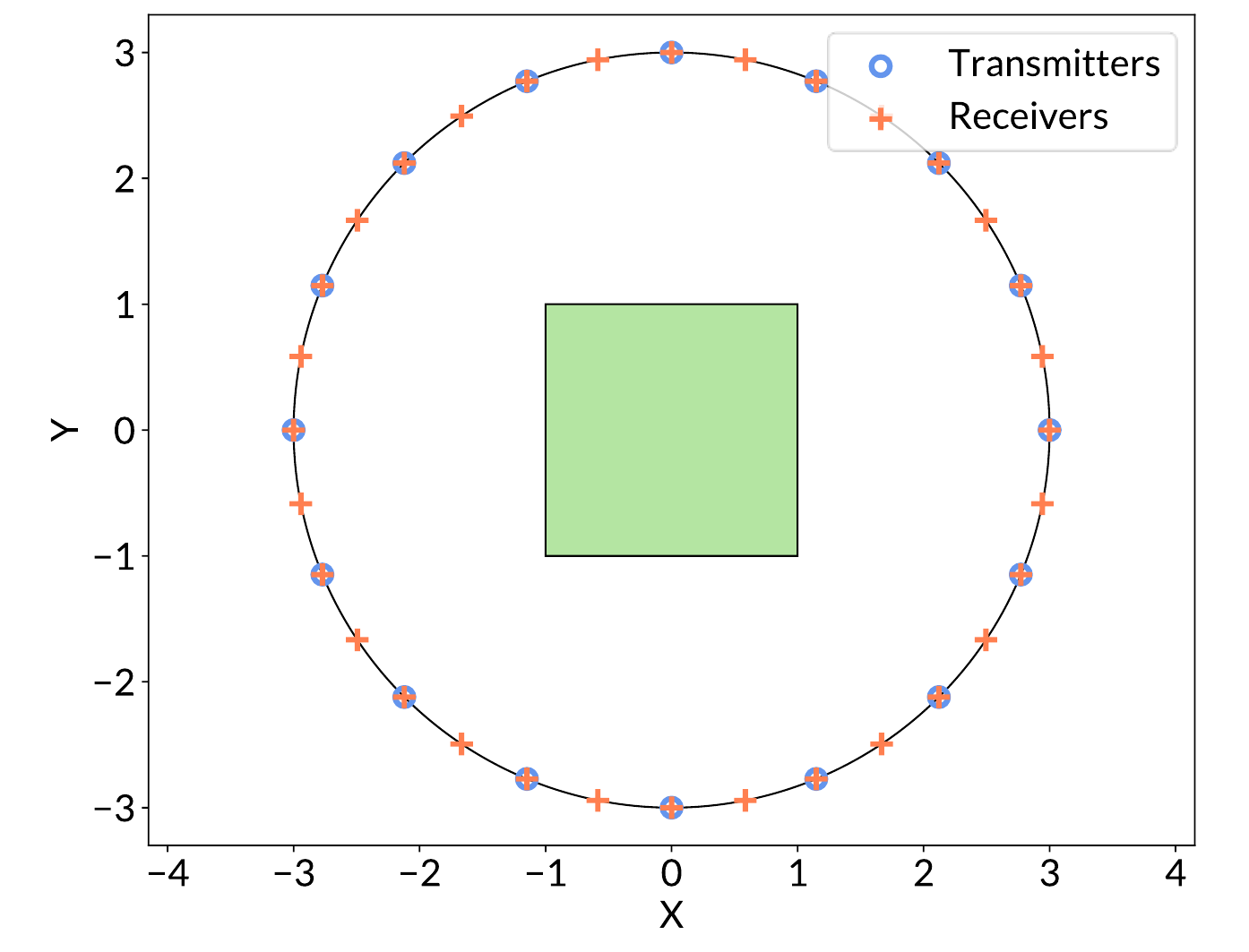}}
    \subfloat[\label{fig:schematic_IF}]{
        \includegraphics[width=0.49\linewidth]{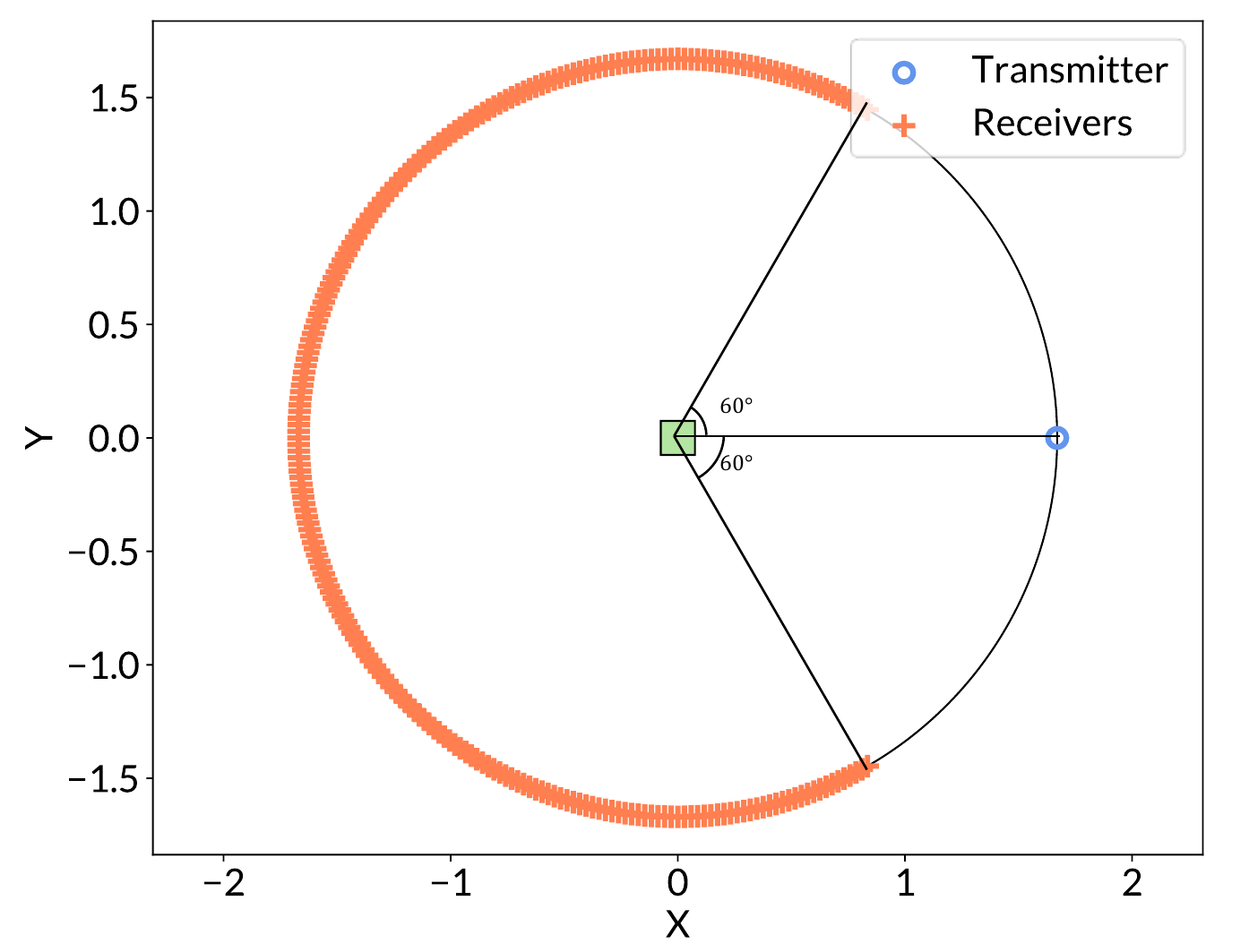}}
\caption{\textbf{Positions of transmitters and receivers for 2D data.} (a) For \ac{circular} dataset and \ac{mnist} dataset, we set up $\Ni=16$ transmitters and $\Nr=32$ receivers. All transmitters and receivers are equally placed. For single-transmitter settings, the transmitter is positioned at the maximum x-coordinate at (3, 0). (b) For \ac{ifdata} dataset, we set up $\Ni=8$ or $18$ transmitters and $\Nr=241$ receivers. The transmitters are equally placed (not shown in the figure), and the locations of receivers are determined by the transmitter. For all datasets, the green square represents the region of interest $\D$. }
    \label{fig:schematic}
    \vspace{-0.3cm}
\end{figure*}

Following previous work \citep{chen2018computational, zhou2022deep} to generate the above two synthetic datasets, we set operating frequency $f = 400$ MHz. The region of interest is a square with the size of $2m\times 2m$. We use $16$ transmitters and $32$ receivers equally placed on a circle with a radius of $3m$ ($\Nr=32$ and $\Ni=16$). The schematic diagram of the locations of the transmitters and receivers around the region of interest is shown in \cref{fig:schematic_synthetic}. The data are generated numerically using the method of moments~\citep{peterson1998computational} with a $224 \times 224$ grid mesh to avoid inverse crime~\citep{colton1998inverse}. To simulate the noise in actual measurement, we add a $5\%$ level of noise to the scattered field $\Esg$ for regular setting.

3) Real-world \ac{ifdata} dataset \citep{2005InvPr} contains three different dielectric scenarios, namely \fde, \fdi, and \ftd. $\Ni=8$ for \fde and \fdi, $\Ni=18$ for \ftd, and $\Nr=241$
for all the cases. The region of interest is a square with the size of $0.15m\times 0.15m$. Transmitters and receivers are placed on a circle with a radius of $1.67m$. The transmitters are placed equally, and the locations of receivers vary for each transmitter. There is no receiver at any position closer than $60^{\circ}$ from the transmitter, and 241 receivers are placed from $+60^{\circ}$ to $+300^{\circ}$ with a step of $1^{\circ}$ from the location of the transmitter. The schematic diagram of the locations of the transmitters and receivers around the region of interest is shown in \cref{fig:schematic_IF}. In the real measurement, there is only one transmitter at a fixed location, and the scatterer rotates to simulate the transmitter to be in different directions. There is a movable receiver sequentially measures the scattered field at 241 different locations. After the measurement, the scatterer rotates by a certain angle for next measurement. The angle is $45^{\circ}$ for \fde and \fdi because $\Nr=8$, and $20^{\circ}$ for \ftd because $\Nr=18$. As for operating frequency, all cases are measured under many different frequencies, and we take the frequency $f = 5$ GHz. We evaluate these three scenarios for testing, and use the same settings to synthetically generate $10k$ images of cylinders with random number and location for training purposes. 

\begin{figure*}
    \centering
    \includegraphics[width=0.6\linewidth]{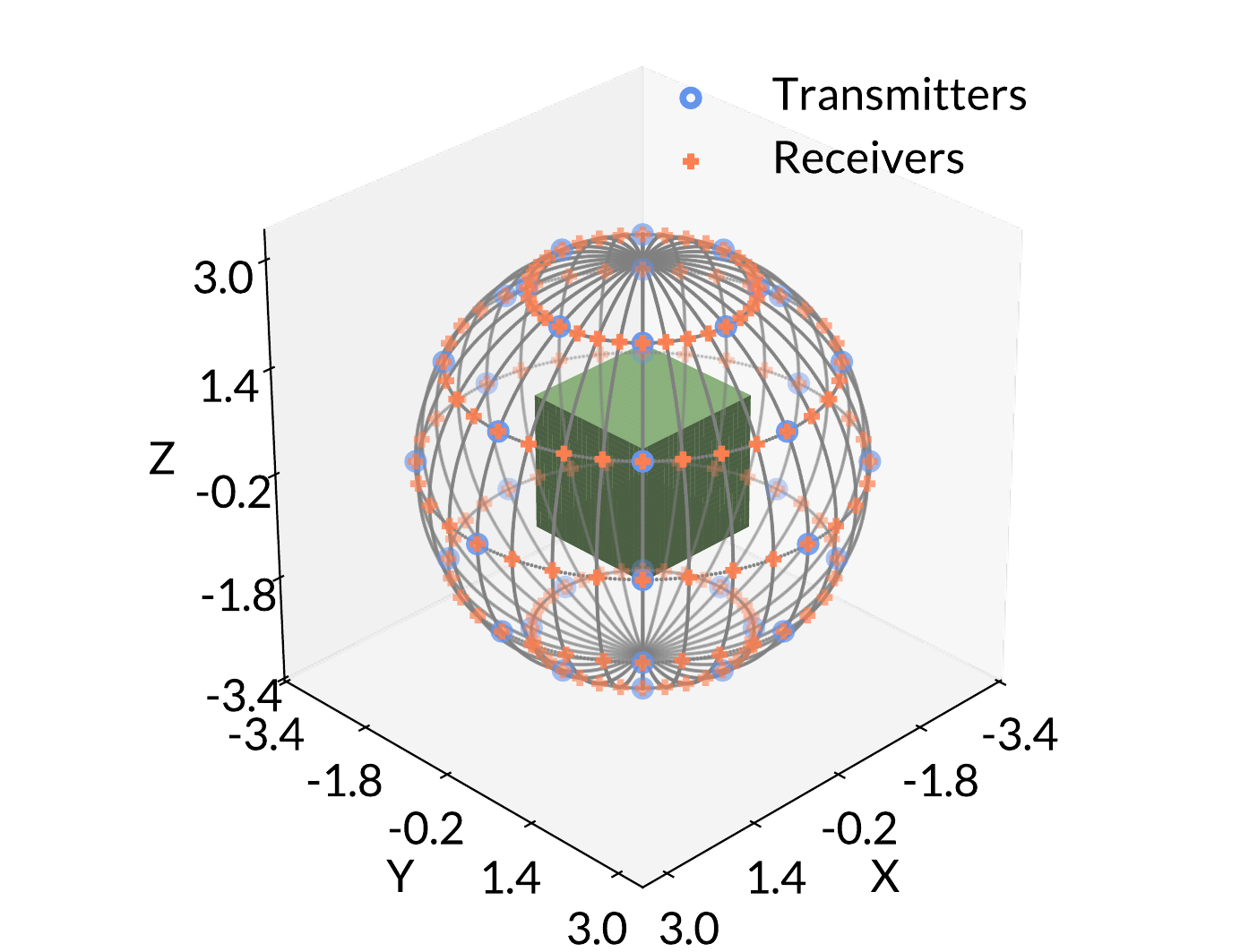}
\caption{\textbf{Positions of transmitters and receivers for \ac{3dmnist} dataset.} There are $\Ni=40$ transmitters and $\Nr=160$ receivers. The green cube represents the region of interest $\D$. }
    \label{fig:schematic_3D}
    \vspace{-0.3cm}
\end{figure*}

4) Synthetic \ac{3dmnist} \citep{3Dmnist} contains 3D images of handwritten digits. 5) Synthetic \ac{3dshapenet} \citep{wu20153d} contains 3D images of various shapes. We use these two datasets to synthesize scatterers with relative permittivity value of $2$. $\Nr=160$ and $\Ni=40$, and the operating frequency $f = 400$ MHz. The region of interest is a cube with the size of $2m\times 2m \times 2m$. The transmitters and receivers are placed at a sphere with the radius of $3m$. For the positions of transmitters, the azimuthal angle ranges from $0^{\circ}$ to $315^{\circ}$ with the step of $45^{\circ}$, and the polar angle ranges from $30^{\circ}$ to $150^{\circ}$ with the step of $30^{\circ}$. For the positions of receivers, the azimuthal angle ranges from $0^{\circ}$ to $348.75^{\circ}$ with the step of $11.25^{\circ}$, and the polar angle ranges from $30^{\circ}$ to $150^{\circ}$ with the step of $30^{\circ}$. The schematic diagram of the locations of the transmitters and receivers around the region of interest is shown in \cref{fig:schematic_3D}. The entire 3D MNIST dataset is used, including $5k$ images for training purposes and $1k$ images for testing. For \ac{3dshapenet}, we take $5k$ images from $5$ different categories for training purposes and $500$ images for testing.

\subsection{Metrics}
Following previous work~\citep{liu2022physics}, we evaluate the quantitative performance of our method using PSNR \citep{wang2009mean}, SSIM \citep{wang2004image}, Relative Root-Mean-Square Error (MSE)~\citep{song2021electromagnetic} and \ac{iou}. For PSNR, SSIM and \ac{iou}, a higher value indicates better performance. For \ac{rrmse}, a lower value indicates better performance. \ac{rrmse} is a metric widely used in \ac{eisp}, which is
defined as follows:
\begin{equation}
\mathrm{MSE}=({\frac{1}{M \times M} \sum_{m=1}^M \sum_{n=1}^M|\frac{\hat{\boldsymbol{\varepsilon}}_r(m, n)-\boldsymbol{\varepsilon}_r(m, n)}{\boldsymbol{\varepsilon}_r(m, n)}|^2})^{\frac{1}{2}},
\end{equation}
where $\boldsymbol{\varepsilon}_r(m, n)$ and $\hat{\boldsymbol{\varepsilon}}_r(m, n)$ are the \ac{gt} and predicted discrete relative permittivity of the unknown scatterers at location $(m, n)$, respectively, and $M \times M$ is the total number of subunits over the \ac{roi} $\D$.

\subsection{Implementation Details}
We implement our method using PyTorch. Our \ac{mlp} consists of 8 layers, each with 512 channels. ReLU activation is used between layers to ensure nonlinear expressiveness. Apply positional encoding to $\x$ before inputting it into \ac{mlp}.

During training, we discretize the region of interest $\D$ into a $64\times64$ grid and optimize the model using the Adam optimizer with default values $\beta_1=0.9, \beta_2=0.999,\epsilon=10^{-8}$, and an initial learning rate of $1\times 10^{-3}$, which remains constant throughout the entire training process. Positional encoding uses frequency $\Omega = 10$, and training runs for 200 iterations on an NVIDIA V100 GPU.

\section{Additional Experimental Results} 
\label{supp:sec:exp}
This section provides additional experimental results, including statistical validation (\cref{supp:sec:exp:errorbar}), extended qualitative comparisons (\cref{supp:sec:exp:qualitative_results}) and additional ablation studies on loss functions (\cref{supp:sec:exp:loss_ablation}). The qualitative comparisons include more visual comparisons with baseline methods, while the ablations further analyze noise robustness, training data size, and the impact of different training losses.

\subsection{Experimental Statistical Significance}
\label{supp:sec:exp:errorbar}
Given the inherent stochasticity of the additive noise in our experiments, all quantitative results reported in the main manuscript represent averages over 3 independent trials. The noise samples were independently drawn from a $\mathcal{N}(0,1)$ distribution and subsequently scaled according to both the signal amplitude and the predefined noise ratios (5\% and 30\%). We report the mean and standard deviation (std) in Table~\ref{tab:errorbar_result}, which shows that the experimental variations introduced less than 1\% error, confirming the statistical significance of our results.

\begin{table*}[h]
\vspace{-7pt}
 \centering
\caption{\textbf{Statistical results of quantitative metrics.} For Circular and MNIST datasets, we report the mean and standard deviation (std) over 3 independent trials under two noise levels: 5\% and 30\%.}
\label{tab:errorbar_result}
\resizebox{\linewidth}{!}{ 
\setlength{\tabcolsep}{0.3mm}
\begin{tabular}{l|ccc|ccc|ccc|ccc}
\thickhline
\multirow{2}{*}{} & \multicolumn{3}{c|}{MNIST (5\%)} & \multicolumn{3}{c|}{MNIST (30\%)} & \multicolumn{3}{c|}{Circular (5\%)} & \multicolumn{3}{c}{Circular (30\%)}  \\
& MSE $\downarrow$         & SSIM $\uparrow$        & PSNR $\uparrow$        & MSE $\downarrow$        & SSIM $\uparrow$         & PSNR $\uparrow$        & MSE $\downarrow$       & SSIM $\uparrow$       & PSNR $\uparrow$       & MSE $\downarrow$       & SSIM $\uparrow$        & PSNR $\uparrow$  \\  
\thickhline
\rowcolor{mygray}
\multicolumn{13}{c}{Number of Transmitters: $N=16$} \\
 mean   &         0.039&        0.978&        32.11&         0.050&         0.966&       29.91  &  0.020&          0.965&         36.92&           0.024&          0.954&         35.19 \\
std & $2.8\times10^{-5}$ & $5.2\times10^{-5}$ & $6.2\times10^{-3}$ & $2.1\times10^{-4}$ & $2.4\times10^{-4}$ & $1.7\times10^{-2}$ & $1.6\times10^{-5}$ & $4.8\times10^{-5}$ & $7.6\times10^{-3}$ & $7.1\times10^{-5}$ & $3.3\times10^{-4}$ & $1.4\times10^{-2}$  \\
\thickhline
\rowcolor{mygray}
\multicolumn{13}{c}{Number of Transmitters: $N=1$} \\
mean & 0.085& 0.921& 26.09& 0.127& 0.862& 22.56& 0.031 & 0.931& 33.18& 0.038 & 0.914& 31.38\\
std & $2.1\times10^{-4}$ & $3.1\times10^{-4}$ & $2.0\times10^{-2}$ & $8.1\times10^{-5}$ & $7.2\times10^{-4}$ & $2.2\times10^{-2}$ & $5.8\times10^{-5}$ & $1.3\times10^{-4}$ & $1.2\times10^{-2}$ & $2.6\times10^{-4}$ & $5.2\times10^{-4}$ & $3.3\times10^{-2}$ \\

    \thickhline
 \end{tabular}}
\end{table*}

\subsection{Additional Qualitative Results}
\label{supp:sec:exp:qualitative_results}
\begin{figure*}[h]
    \centering
    \includegraphics[width=.9\linewidth]{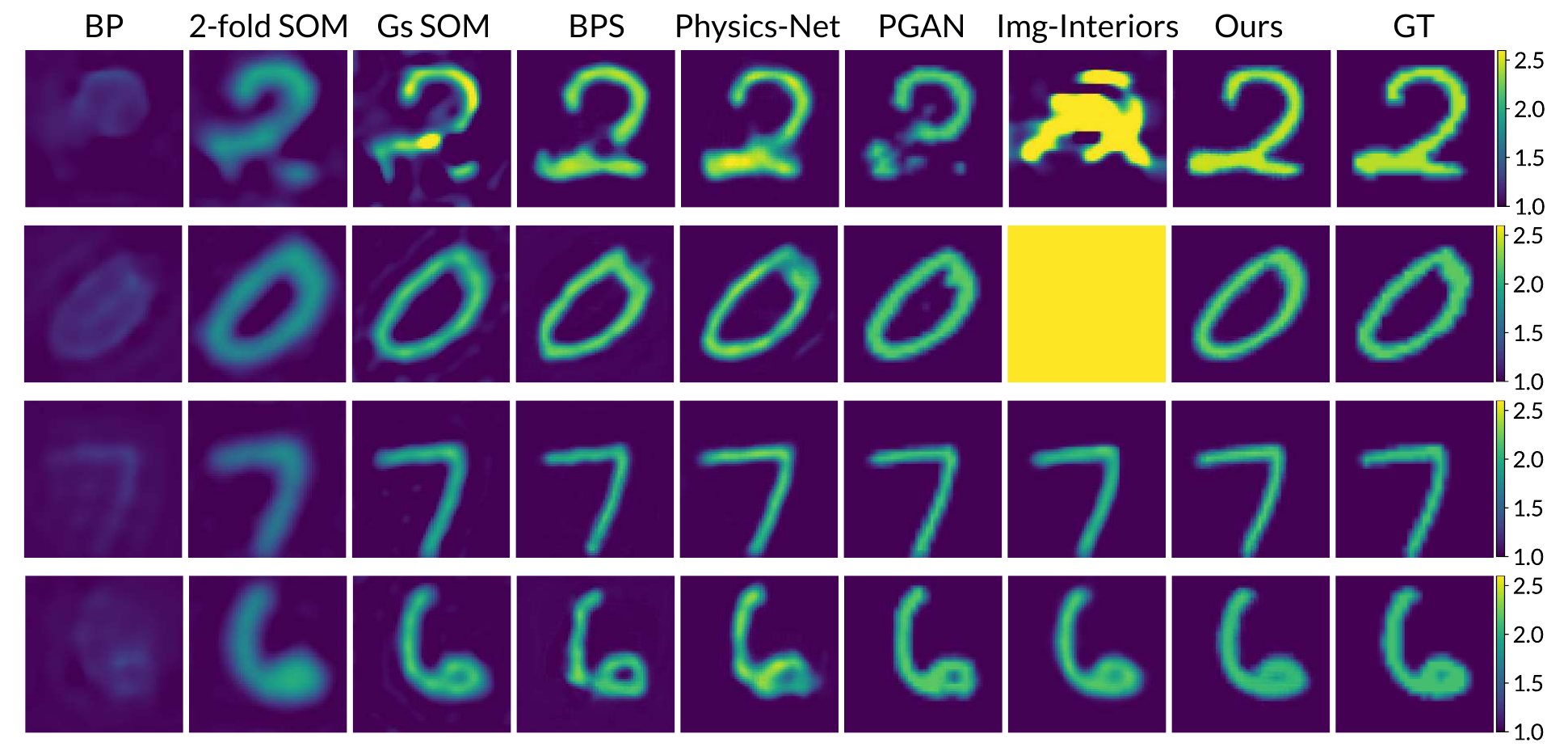}
    \caption{\textbf{Qualitative comparison under the multiple-transmitter setting on \ac{mnist} dataset.} The results are obtained with $\Ni=16$ transmitters and a noise level of 5\%. Colors represent the values of the relative permittivity.}
    \label{fig:MNIST_0.05noise_N_inc=16_supp}
    \vspace{-0.2cm}
\end{figure*}

\begin{figure*}[h]
    \centering
    \includegraphics[width=.9\linewidth]{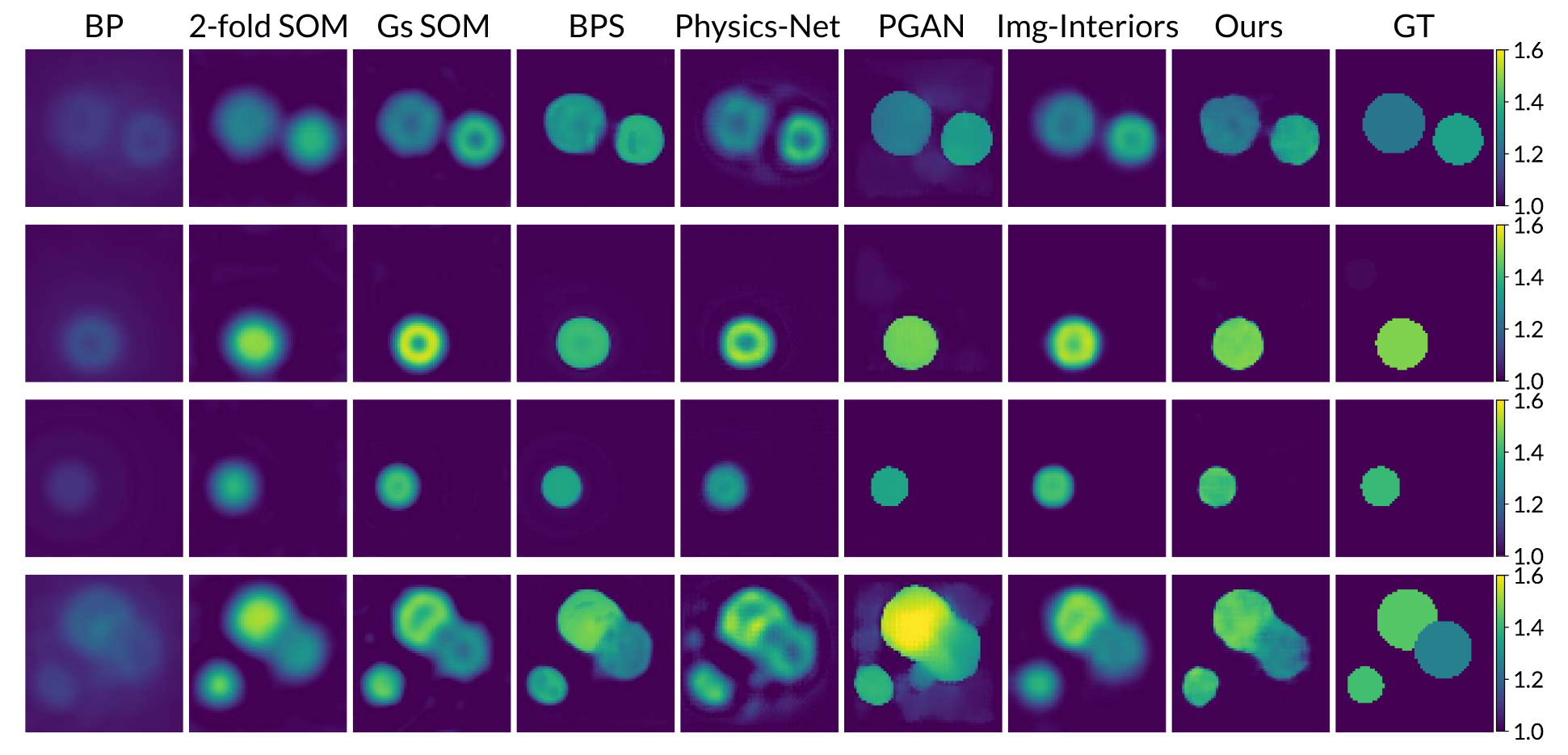}
    \caption{\textbf{Qualitative comparison under the multiple-transmitter setting on \ac{circular} dataset.} The results are obtained with $\Ni=16$ transmitters and a noise level of 5\%. Colors represent the values of the relative permittivity.}
    \label{fig:cylinder_0.05noise_N_inc=16_supp}
    \vspace{-0.5cm}
\end{figure*}

\begin{figure*}[h]
    \centering
    \includegraphics[width=.9\linewidth]{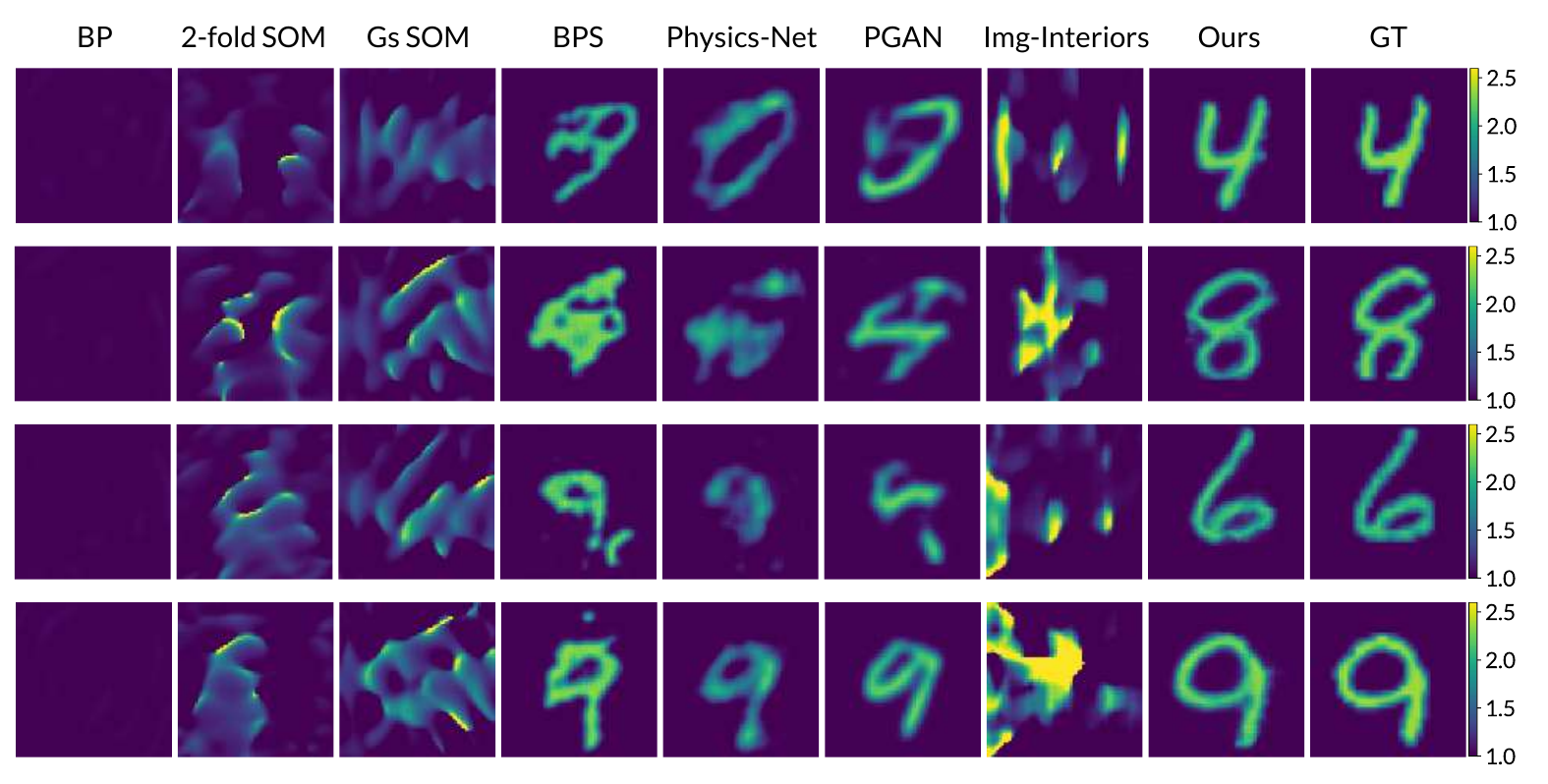}
    \caption{\textbf{Qualitative comparison under the single-transmitter setting on \ac{mnist} dataset.} The results are obtained with $\Ni=1$ transmitter and a noise level of 5\%. Colors represent the values of the relative permittivity.}
    \label{fig:MNIST_0.05noise_N_inc=1_supp}
    \vspace{-0.5cm}
\end{figure*}

\begin{figure*}[h]
    \centering
    \includegraphics[width=.9\linewidth]{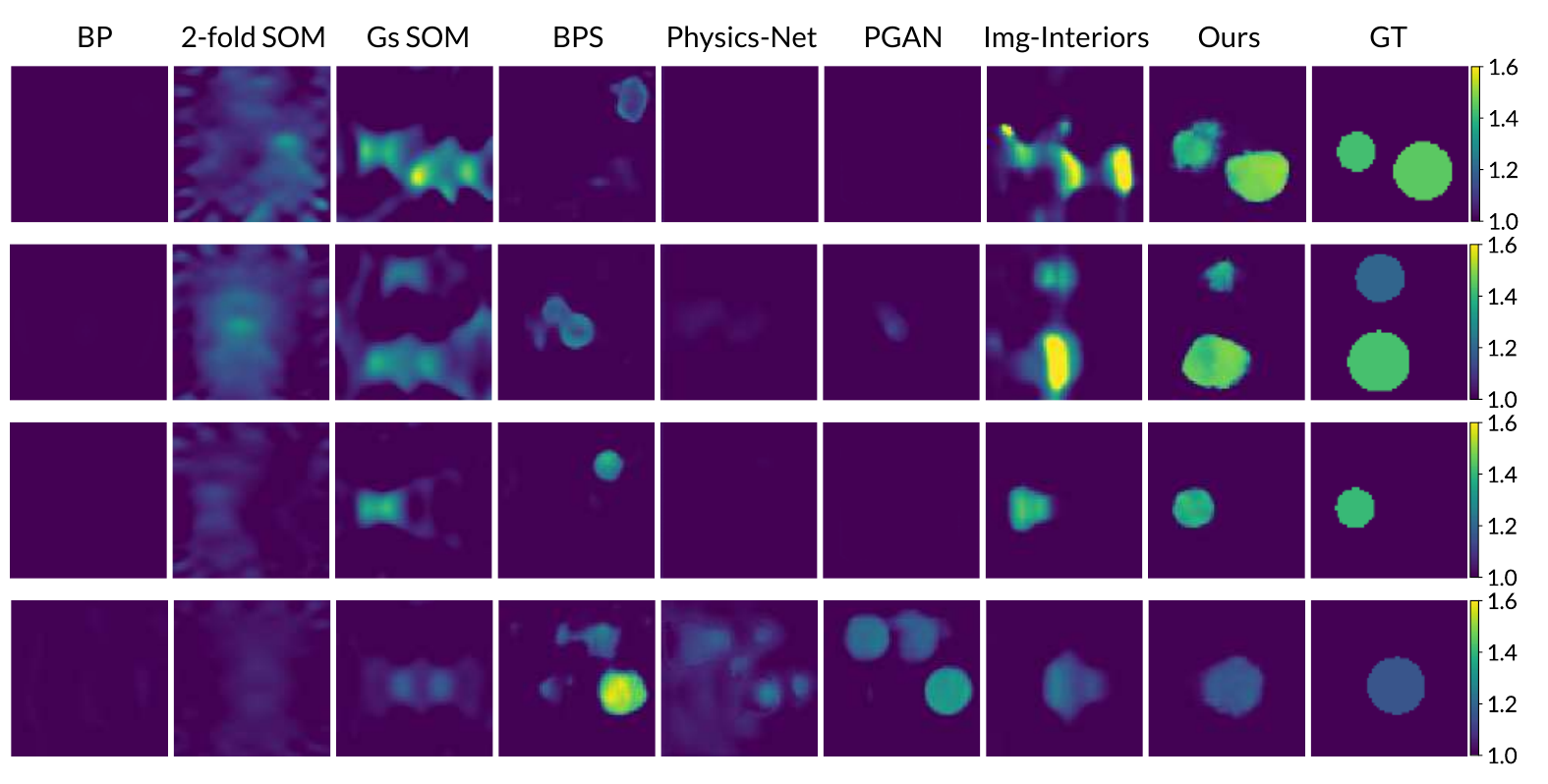}
    \caption{\textbf{Qualitative comparison under the single-transmitter setting on \ac{circular} dataset.} The results are obtained with $\Ni=1$ transmitter and a noise level of 5\%. Colors represent the values of the relative permittivity.}
    \label{fig:cylinder_0.05noise_N_inc=1_supp}
    \vspace{-0.5cm}
\end{figure*}

\begin{figure*}[h]
    \centering
    \includegraphics[width=.9\linewidth]{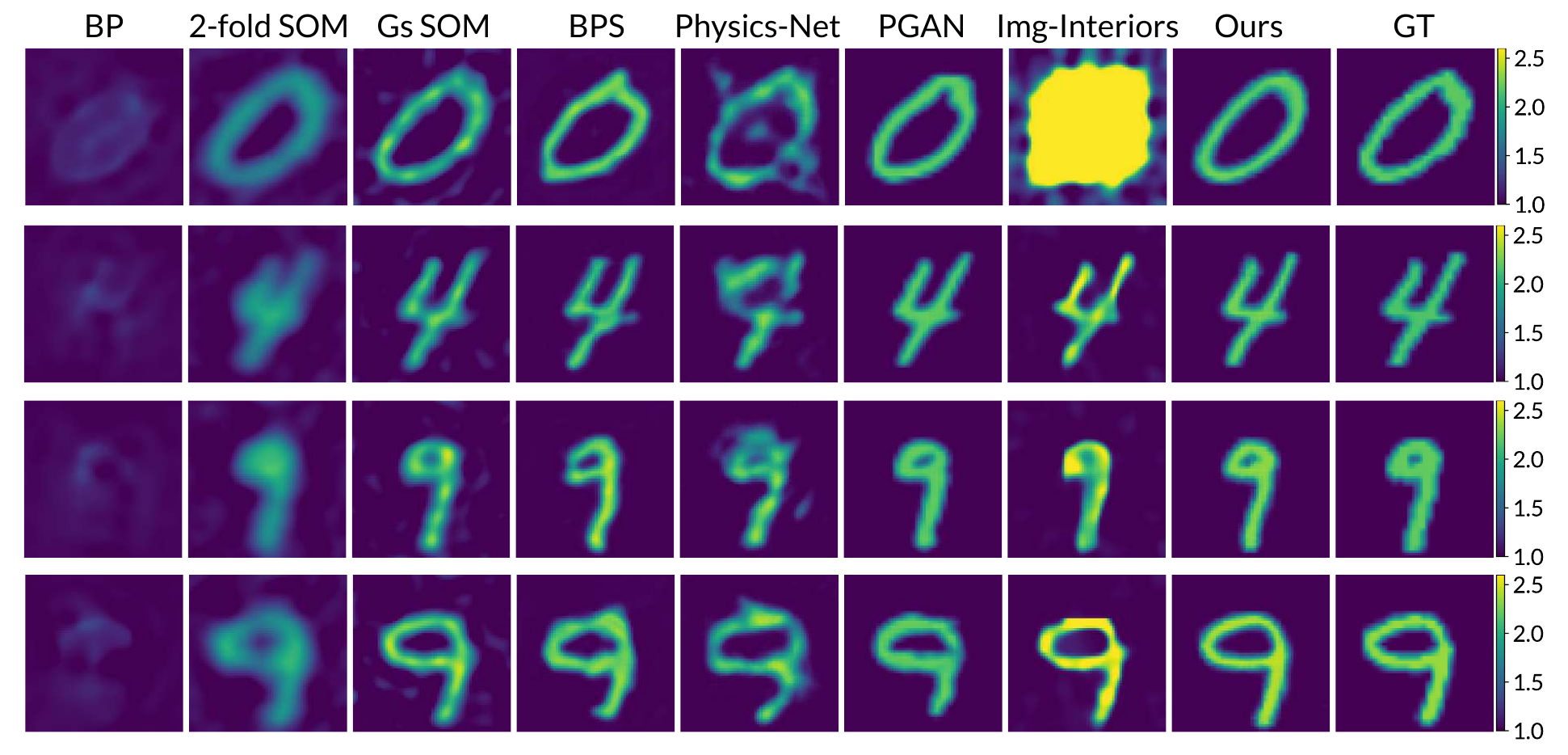}
        \caption{\textbf{Qualitative comparison under high noise setting on \ac{mnist} dataset.} The results are obtained with $\Ni=16$ transmitters and a noise level of 30\%. Colors represent the values of the relative permittivity.}
    \label{fig:MNIST_0.30noise_N_inc=16_supp}
    \vspace{-0.2cm}
\end{figure*}

\begin{figure*}[h]
    \centering
    \includegraphics[width=.9\linewidth]{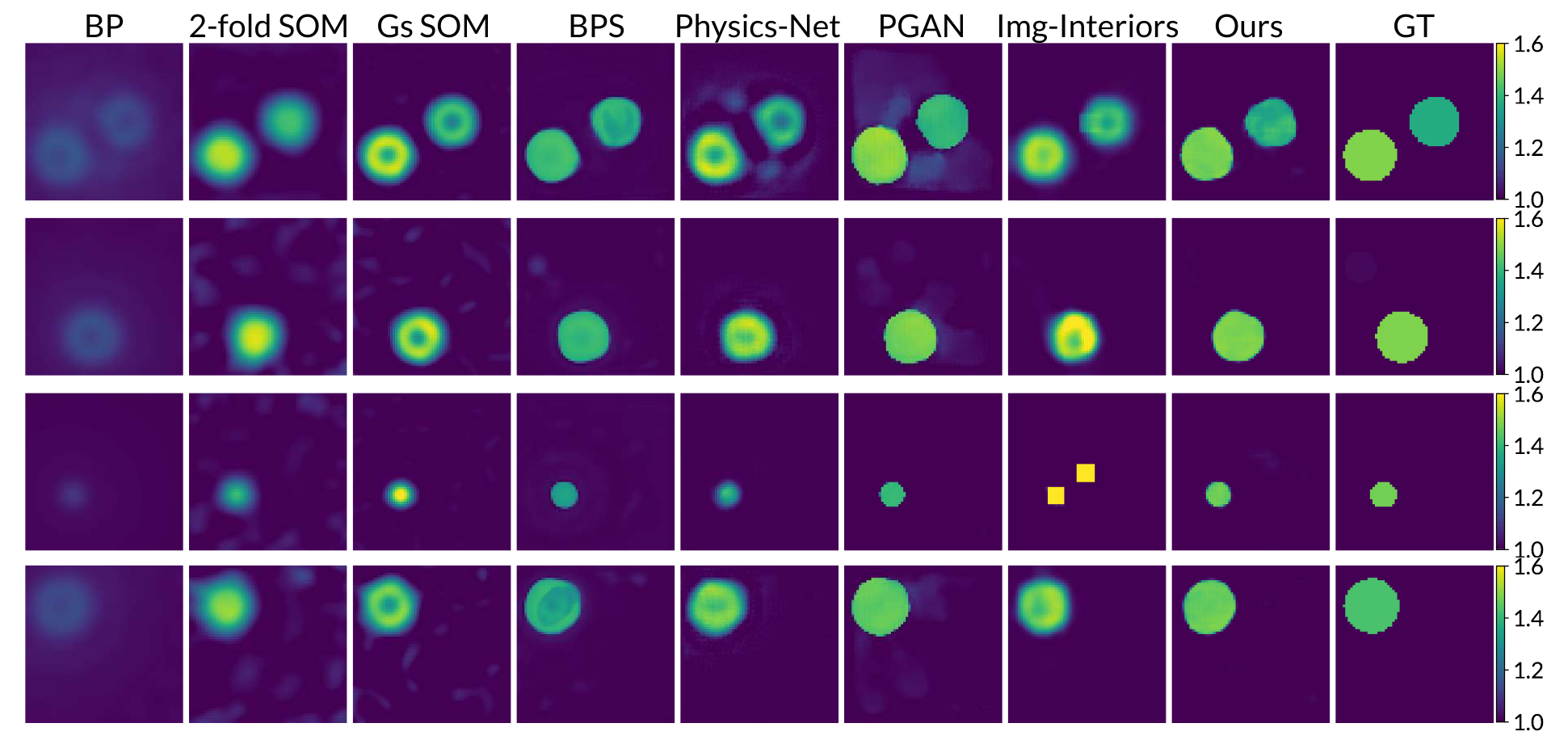}
        \caption{\textbf{Qualitative comparison under high noise setting on \ac{circular} dataset.} The results are obtained with $\Ni=16$ transmitters and a noise level of 30\%. Colors represent the values of the relative permittivity.}
    \label{fig:cylinder_0.30noise_N_inc=16_supp}
    \vspace{-0.2cm}
\end{figure*}

\paragraph{Qualitative Comparison.} We present more qualitative comparison on \ac{circular} dataset \citep{luo2024imaging} and \ac{mnist} dataset \citep{deng2012mnist} under settings with different transmitter numbers $\Ni$ and noise levels, as shown in \cref{fig:MNIST_0.05noise_N_inc=16_supp} to \cref{fig:cylinder_0.30noise_N_inc=16_supp}. Our method achieves comparable or superior performance to \ac{sota} methods in most cases under multiple-transmitter settings, as \cref{fig:MNIST_0.05noise_N_inc=16_supp}, \cref{fig:cylinder_0.05noise_N_inc=16_supp}, \cref{fig:MNIST_0.30noise_N_inc=16_supp}, and \cref{fig:cylinder_0.30noise_N_inc=16_supp} indicate. As shown in \cref{fig:cylinder_0.05noise_N_inc=16_supp} and \cref{fig:cylinder_0.30noise_N_inc=16_supp}, \pg \citep{song2021electromagnetic} introduces noticeable background artifacts due to the lack of consideration of physics. And \cref{fig:MNIST_0.05noise_N_inc=16_supp} shows that \iishort occasionally fails to converge due to local optima. 

For single-transmitter setting ($\Ni=1$), our method significantly outperforms all previous approaches across all datasets, as shown in \cref{fig:MNIST_0.05noise_N_inc=1_supp} and \cref{fig:cylinder_0.05noise_N_inc=1_supp}. Conventional methods such as \ac{bp}\citep{belkebir2005superresolution}, \gs\citep{chen2009subspace}, and \tfshort~\citep{chen2009subspace} produce only blurry reconstructions. Deep learning-based methods such as \bps~\citep{chen2018computational}, \pn~\citep{liu2022physics}, and \pg~\citep{song2021electromagnetic} tend to ``hallucinate'' the digit, leading to wrong reconstruction on the \ac{mnist} dataset. \iishort~\citep{luo2024imaging} produces results with structural errors that deviate significantly from the true morphology. In contrast, our method can still produce reasonably accurate reconstructions of the relative permittivity under such a challenging condition. This enables practical applications with fewer transmitters, significantly reducing deployment costs while preserving reconstruction quality. 

\begin{figure*}[h]
    \centering
    \includegraphics[width=.9\linewidth]{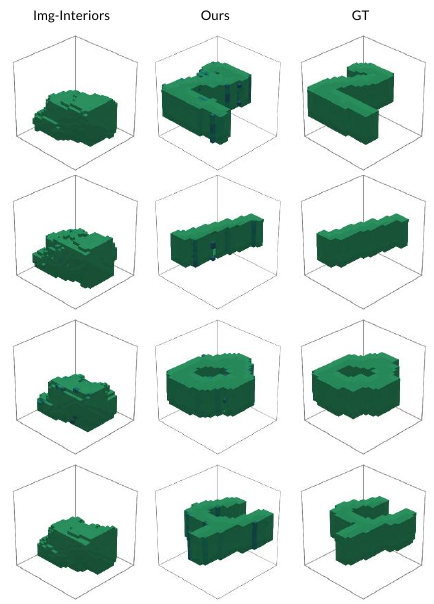}
    \caption{\textbf{Qualitative comparison under the single-transmitter setting for 3D reconstruction on \ac{3dmnist} dataset.} The results are obtained with single transmitter ($\Ni = 1$). The voxel colors represent the values of the relative permittivity.}
    \label{fig:3Dfigsupp}
\end{figure*}

\paragraph{3D Reconstruction.} We present more qualitative comparison on \ac{3dmnist}~\citep{3Dmnist} dataset under the single-transmitter setting, as shown in \cref{fig:3Dfigsupp}. Our method successfully approximates permittivity reconstruction even in this difficult setting, whereas \iishort \citep{luo2024imaging} fails. Similar to the 2D scenario, \iishort suffers from degeneration and can only produce fixed patterns, failing to capture the shape of scatterers. 

\begin{figure*}[h]
    \centering
    \includegraphics[width=.9\linewidth]{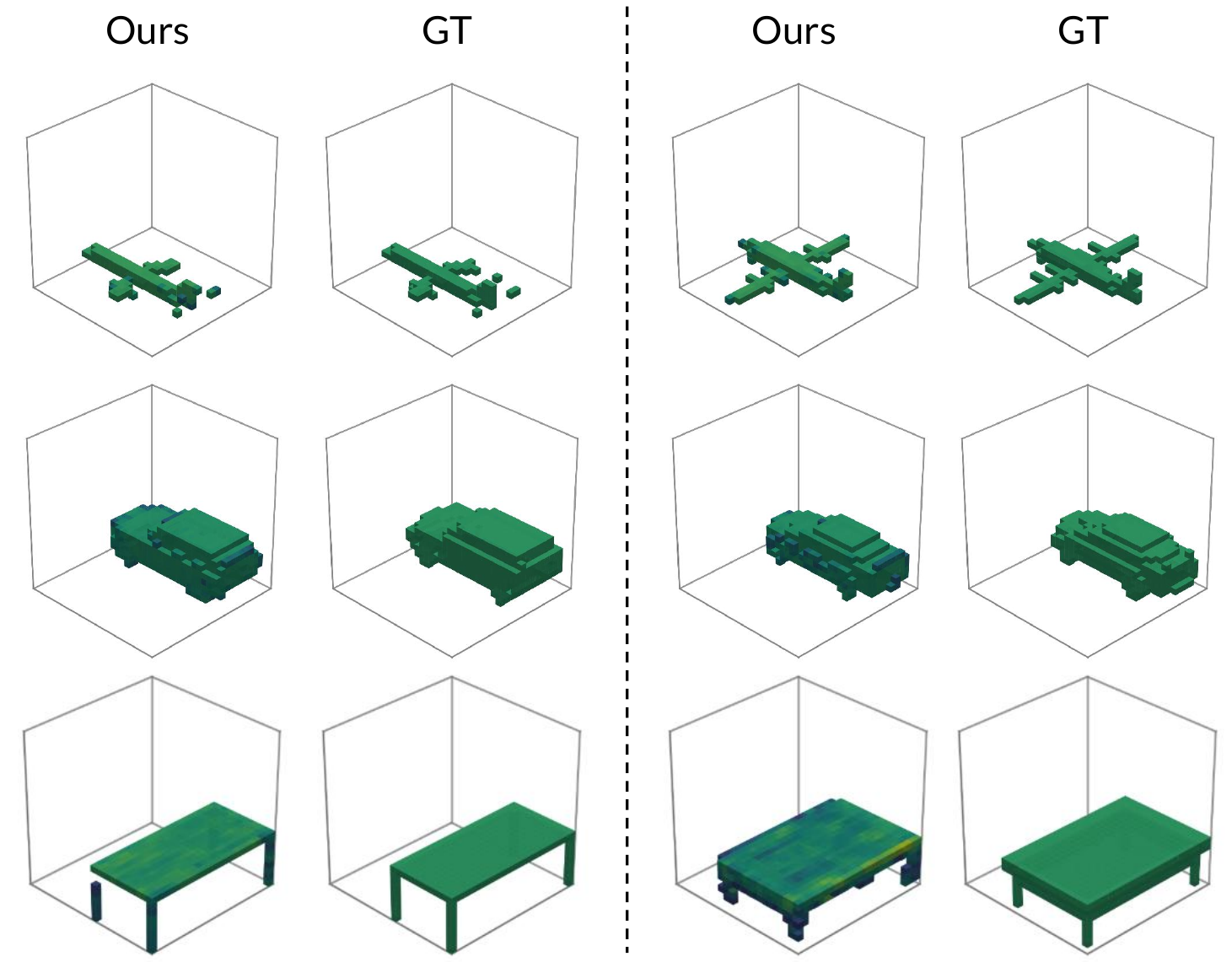}
    \caption{\textbf{Qualitative comparison under the single-transmitter setting for 3D reconstruction on \ac{3dshapenet} dataset.} The results are obtained with single transmitter ($\Ni = 1$). The voxel colors represent the values of the relative permittivity.}
    \label{fig:3Dshapenetfigsupp}
\end{figure*}

Additionally, we evaluate our method on the more complex \ac{3dshapenet}~\citep{wu20153d} dataset to demonstrate its generalization capability. As shown in \cref{fig:3Dshapenetfigsupp}, our approach can successfully reconstruct various complex structures including airplanes, cars and tables under the same single-transmitter configuration, maintaining accurate permittivity distribution recovery. This demonstrates the strong potential of our method for practical real-world applications.

\begin{figure*}[h]
    \centering
    \includegraphics[width=.85\linewidth]{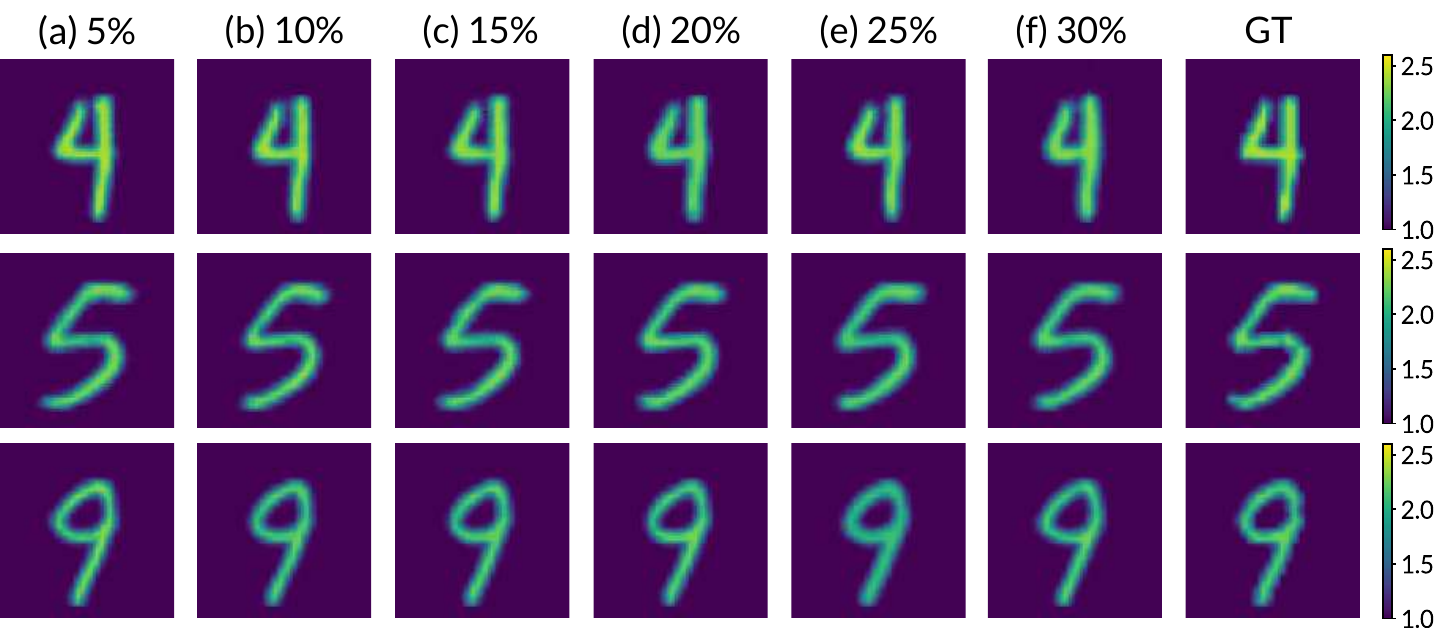}
    \caption{\textbf{Qualitative results of our noise robustness on the \ac{mnist} dataset.} The results are
obtained with $\Ni = 16$ transmitter. Colors represent the values of the relative permittivity.
}
    \vspace{-0.8cm}
    \label{fig:transmitter_ablation}
\end{figure*}

\paragraph{Noise Robustness.} We present qualitative results across multiple noise levels (ranging from 5\% to 30\%) on the \ac{mnist} \citep{deng2012mnist} dataset, providing visual support for the noise robustness analysis. As shown in \cref{fig:transmitter_ablation}, our method maintains high visual fidelity at low noise levels. More importantly, the method maintains stable reconstructions and preserves the essential structure of the scatterer across the entire tested noise range (5\% to 30\%), demonstrating a smooth and gradual degradation in quality that is fully consistent with the quantitative results.

\begin{figure*}[h]
    \centering
    \includegraphics[width=.72\linewidth]{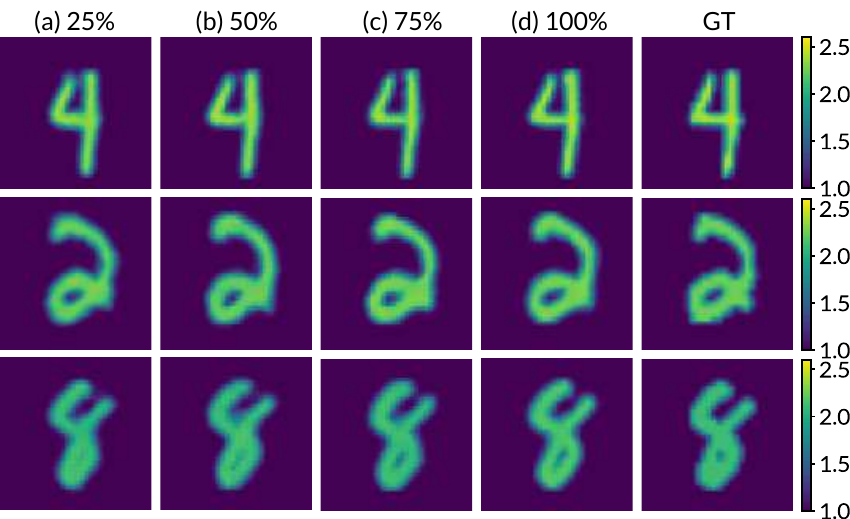}
    \caption{\textbf{Qualitative results of training data size ablation on the \ac{mnist} dataset.} The results are obtained with $\Ni = 16$ transmitters and a noise level of 5\%.}
    \vspace{-0.8cm}
    \label{fig:ablation_datasize05}
\end{figure*}

\begin{figure*}[h]
    \centering
    \includegraphics[width=.72\linewidth]{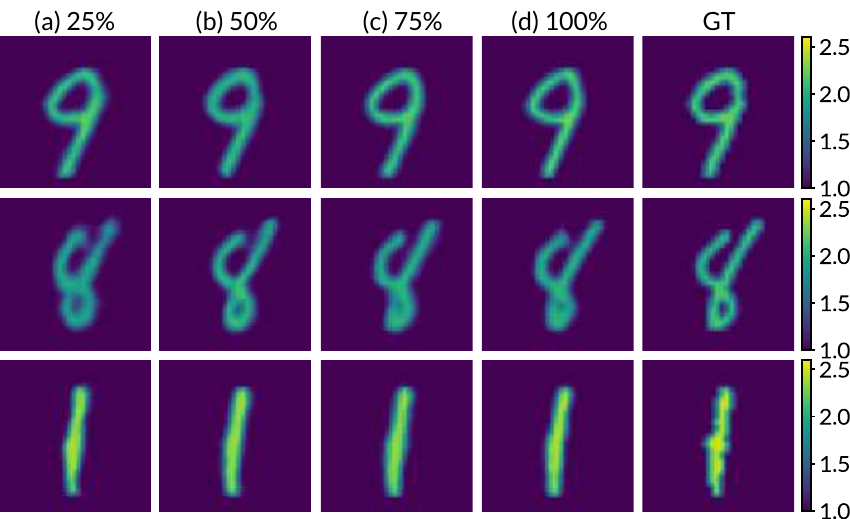}
    \caption{\textbf{Qualitative results of training data size ablation on the \ac{mnist} dataset.} The results are obtained with $\Ni = 16$ transmitters and a noise level of 30\%.}
    \vspace{-0.8cm}
    \label{fig:ablation_datasize30}
\end{figure*}

\paragraph{Ablation on Training Data Size.}  We present qualitative results across different training data scales (ranging from 25\% to 100\%) on the \ac{mnist} \citep{deng2012mnist} dataset, offering visual insights into the impact of data volume on reconstruction performance. As shown in \cref{fig:ablation_datasize05} and \cref{fig:ablation_datasize30}
, our model maintains satisfactory reconstruction integrity even with limited training data. 

\subsection{Additional Ablation Results}
\label{supp:sec:exp:loss_ablation}
We experiment with TV loss~\cite{rudin1992} for smoothness and Perceptual loss~\cite{johnson2017googlesmultilingualneuralmachine} for structure in addition to MSE on single-transmitter setting. \cref{tab:ablation_loss} reports the quantitative results
on the MNIST dataset under different noise levels. Overall, the auxiliary losses provide mixed improvements, while MSE alone already provides strong and stable
performance, indicating that our model is relatively robust
to the choice of loss function.
\begin{table}[h]
\centering
\small
\caption{\textbf{Ablation of loss functions.} 
We compare MSE with auxiliary TV loss~\cite{rudin1992} and perceptual loss~\cite{johnson2017googlesmultilingualneuralmachine} under the single-transmitter setting. 
Results are reported on the MNIST dataset under two noise levels: 5\% and 30\%.}
\label{tab:ablation_loss}
\resizebox{1.0\linewidth}{!}{
\begin{tabular}{l|ccc|ccc}
\toprule
\multirow{2}{*}{Loss} 
& \multicolumn{3}{c|}{MNIST (5\%)} 
& \multicolumn{3}{c}{MNIST (30\%)} \\
& MSE $\downarrow$ & SSIM $\uparrow$ & PSNR $\uparrow$
  & MSE $\downarrow$ & SSIM $\uparrow$ & PSNR $\uparrow$ \\
\midrule
Ours (MSE)              & 0.085 & 0.921 & 26.09 & 0.127 & 0.862 & 22.56 \\
Ours (w/ TV)       & 0.084 & 0.917 & 26.00 & 0.124 & 0.859 & 22.74 \\
Ours (w/ Percpt.)    & 0.084 & 0.923 & 26.20 & 0.130 & 0.861 & 22.49 \\
\bottomrule
\end{tabular}}
\vspace{-1.0em}
\end{table}

\clearpage

\end{document}